\documentclass{article} 
\usepackage[final]{neurips_2025} 

\PassOptionsToPackage{table}{xcolor}
\PassOptionsToPackage{hyphens}{url}

\usepackage{booktabs}
\usepackage{longtable}
\usepackage{array}
\usepackage{colortbl}
\usepackage{multicol}
\usepackage{multirow}
\usepackage{threeparttable}
\usepackage{graphicx}
\usepackage{xcolor}

\definecolor{topgreen}{HTML}{D8F0B7}

\newlength{\captionskipamount}  
\newlength{\floatskipamount}    
\usepackage{caption}
\usepackage{subcaption}

\usepackage{listings}
\usepackage{xcolor}

\lstdefinelanguage{json}{
    basicstyle=\ttfamily\small,
    numbers=left,
    numberstyle=\tiny\color{gray},
    stepnumber=1,
    numbersep=6pt,
    showstringspaces=false,
    breaklines=true,
    frame=single,
    backgroundcolor=\color{gray!5},
}

\usepackage[utf8]{inputenc}
\usepackage[T1]{fontenc}

\usepackage[sort,compress]{cite}

\usepackage{hyperref}
\hypersetup{
  colorlinks=true,
  linkcolor=red,
  citecolor=cyan,
  filecolor=magenta,
  urlcolor=blue,
}
\usepackage{url}
\usepackage{amsfonts}
\usepackage{amsmath}
\usepackage{amssymb}
\usepackage{nicefrac}

\renewcommand{\footnoterule}{%
  \kern-3pt%
  \hrule width 0.4\columnwidth height 0.4pt%
  \kern 2.6pt}

\usepackage{microtype}
\usepackage{xspace}
\usepackage{enumitem}
\usepackage{textcomp}
\usepackage{stfloats}
\usepackage{verbatim}
\usepackage{wrapfig}
\usepackage{graphicx}

\usepackage{tikz}
\usepackage{algorithm}
\usepackage{algpseudocode}
\usepackage{pgfplots}
\pgfplotsset{compat=1.18}

\usetikzlibrary{positioning}

\usepackage{placeins}
\newcommand{\matrixicon}{%
\raisebox{-0.03em}{%
  \includegraphics[height=1.0em]{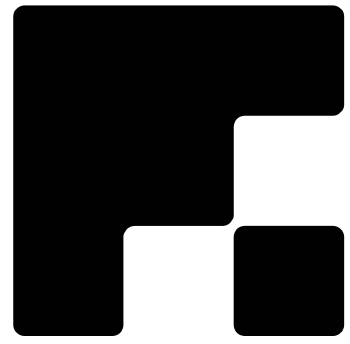}%
}%
\hspace{0.3em}%
}

\title{
\texorpdfstring{
  \matrixicon Matrix-Game 3.5: Enhancing Real-Time Streaming
  Interactive World Models with Patch Memory
}{
  Matrix-Game 3.5: Enhancing Real-Time Streaming
  Interactive World Models with Patch Memory
}
}

\author{
Runjia Qian\thanks{Equal Contribution.}\quad Zile Wang{$^*$}\quad Jihai Zhang{$^*$}\quad Kai Zou{$^*$}\quad Wei Yu{$^*$}\quad Jiaxing Li{$^*$}\quad \\
\textbf{Zexiang Liu{$^*$}\quad Yaokun Li\quad Fei Kang\quad Kaichen Huang\quad Mengyin An\quad }\\
\textbf{Haobo Zhang\quad Biao Jiang\quad Jiahua Wang\quad Haofeng Sun\quad Yang Liu{$^\dagger$}\quad Yangguang Li\thanks{Project Lead and Corresponding Author.  ~\\\\ \newline Technical Report.}}
\\
\\
  Riemann Dynamics \\
   \texttt{research@riemanndynamics.ai}\\
  Project page: \href{https://matrix-game-v3-5.github.io/}{\textcolor{blue}{Matrix-Game-3.5-Homepage}} \quad Code: \href{https://github.com/Riemann-Dynamics/Matrix-Game-3.5}{\textcolor{blue}{GitHub}}\\
  Checkpoints: \href{https://huggingface.co/RiemannDynamics/Matrix-Game-3.5-Base}{\textcolor{blue}{Matrix-Game-3.5-Base}} $|$ \href{https://huggingface.co/RiemannDynamics/Matrix-Game-3.5-Distilled}{\textcolor{blue}{Matrix-Game-3.5-Distilled}}
}

\begin{document}
\maketitle

\begin{abstract}
Interactive world models extend video generation from offline clip synthesis toward persistent simulation of interactive virtual worlds, enabling applications in games, robotics, embodied agents, and XR. Achieving stable long-horizon interactive generation, however, remains challenging, as the model must simultaneously preserve scene geometry, dynamic consistency, and camera control while supporting real-time autoregressive generation.
Building upon Matrix-Game 3.0, we present \textbf{Matrix-Game 3.5}, as shown in Figure~\ref{fig:teaser}, which advances real-time interactive world generation toward geometry-aware and long-horizon consistent simulation through three key improvements. 
First, we propose a unified geometry-aware memory framework, whose patch-memory and tiled-PRoPE components introduce no additional learnable parameters, combining explicit 3D patch retrieval with projective camera conditioning to enable geometry-consistent camera control and faithful long-horizon scene recall. Second, we introduce a static-dynamic disentangled world representation that separately models static scene geometry and dynamic subjects, preserving both geometric consistency and subject identity throughout long-horizon generation. Third, we develop a two-stage progressive real-time distillation framework that converts a bidirectional diffusion model into a few-step causal generator through Perceptual Flow Matching and curriculum-based Self-Rollout DMD, enabling minute-long real-time interactive generation.
Extensive experiments demonstrate that, with a unified training corpus spanning Unreal simulation environments, open-world games, and internet videos, Matrix-Game 3.5 achieves strong performance in long-horizon scene recall, precise camera control, subject consistency, prompt-driven world generation, and stable real-time open-world interaction.
\end{abstract}

\begin{figure}[t!]
    \centering
    \includegraphics[width=1\linewidth]{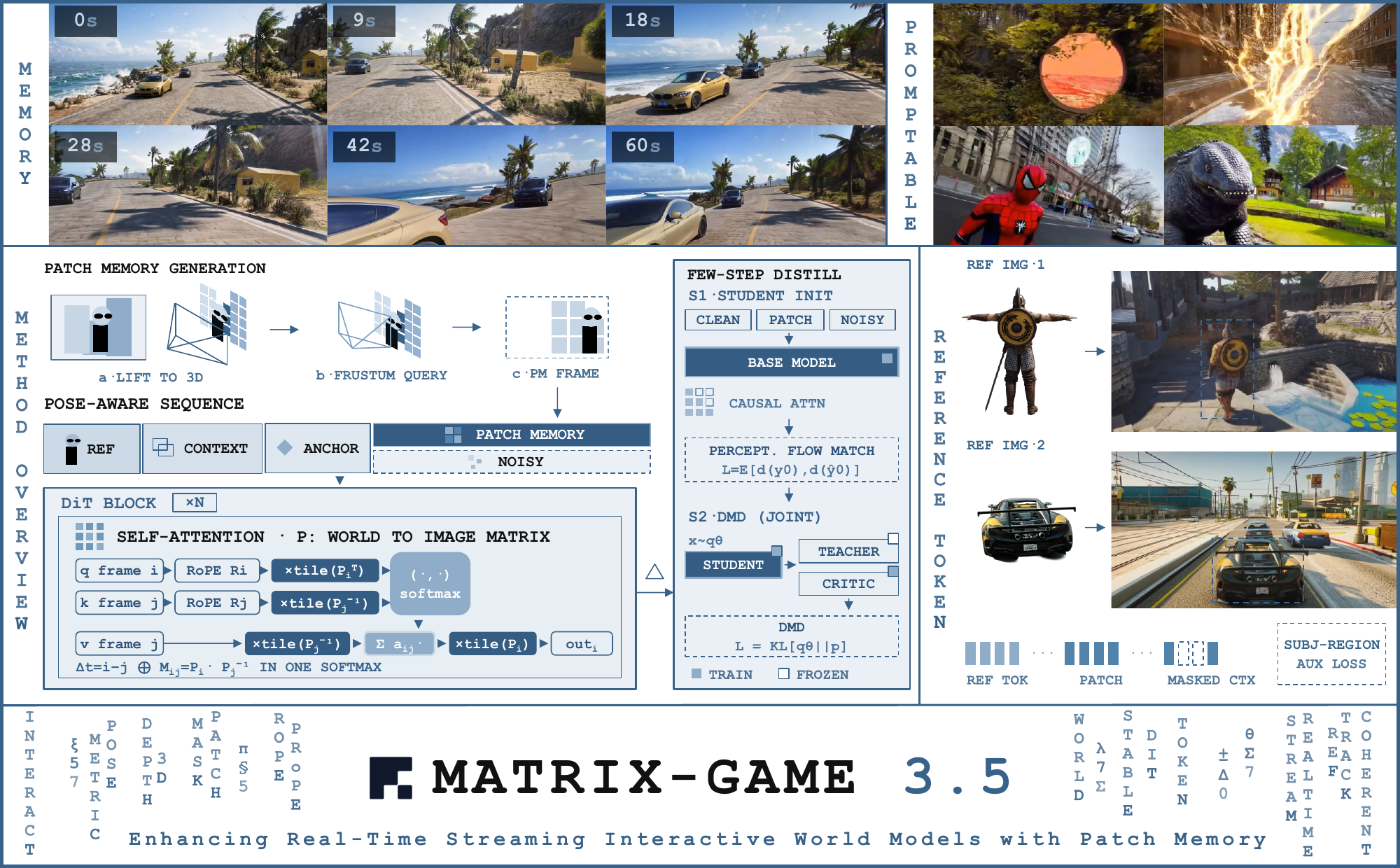}
    \caption{Overview of \textbf{Matrix-Game 3.5}. A real-time interactive world model for long-horizon generation with pose-aware representation, unified memory, and two-stage progressive distillation, achieving single-GPU 720p real-time generation at up to 20 FPS.}
    \label{fig:teaser}
\end{figure}

\section{Introduction}
Interactive world models aim to build persistent virtual environments that continuously evolve in response to user interactions. Unlike conventional video generation models that synthesize short offline clips, interactive world models must causally generate future observations while responding to camera motions, user actions, and high-level prompts. More importantly, the generated world should remain persistent throughout long-horizon interaction: previously visited scenes should be faithfully reconstructed, objects and characters should preserve their identities under viewpoint changes, and user-triggered events should produce coherent and lasting changes to the environment. These capabilities are fundamental for open-world gaming, embodied-agent simulation, robotics, extended reality (XR), and interactive content creation.

Building such persistent worlds remains fundamentally challenging. Interactive world models operate in a closed-loop autoregressive manner, where future predictions continuously become the context for subsequent generation. Consequently, visual and geometric errors inevitably accumulate over time, gradually causing scene drift, identity inconsistency, and degraded controllability. At the same time, the limited context window prevents the model from directly accessing distant observations, making long-term scene revisitation increasingly difficult. Large viewpoint changes further require the model not only to retrieve relevant historical information, but also to determine its correct geometric correspondence under the current camera. Moreover, while static environments should remain persistent over long horizons, dynamic entities are expected to evolve according to user interactions and prompts. Therefore, achieving long-horizon consistency is fundamentally a joint problem of memory, geometry, dynamics, and efficient autoregressive generation.

Existing approaches differ mainly in the unit at which historical observations are stored. One line of work compresses history into latent tokens or transformer memories (e.g., RELIC~\cite{hong2025relic}), enabling efficient long-context modeling while representing historical information only implicitly. A second line stores whole frames and retrieves them through camera geometry, either by field-of-view overlap (Context as Memory~\cite{yu2025context}, WorldMem~\cite{xiao2025worldmem}, Matrix-Game 3.0~\cite{wang2026matrixgame3}) or through reconstructed surfaces that index the stored views (VMem~\cite{li2025vmem}), improving scene revisitation by leveraging explicit spatial cues; a frame, however, remains an indivisible memory unit that can only be retrieved in its entirety and from its original viewpoint. A third line reconstructs the scene explicitly as points or splats, which imposes strong geometric constraints but offers limited flexibility for synthesizing unobserved or newly emerging content. Although these paradigms significantly improve long-horizon consistency, none fully resolves a more fundamental question: \textbf{an interactive world model must jointly reason about what historical information should be remembered, where it should reappear under a novel viewpoint, and whether the remembered content should remain persistent or evolve with the world.} As a result, long-term memory, camera control, dynamic-content consistency, and prompt-driven interaction are still largely addressed as separate problems rather than within a unified world representation.

In this work, we present \textbf{Matrix-Game 3.5}, the latest generation of the Matrix-Game series for real-time long-horizon interactive world generation. Our central idea is to organize historical observations as an explicit geometry-aware persistent memory, maintained at the granularity of image patches rather than whole frames or compressed latent states. Based on this perspective, we introduce a unified geometry-aware memory framework --- whose patch-memory and tiled-PRoPE components introduce no additional learnable parameters --- that integrates \textbf{patch memory}~\cite{yu2026mosaicmem} and \textbf{tiled PRoPE}, our video adaptation of PRoPE~\cite{li2025prope}. The patch memory lifts historical image patches into a persistent 3D memory and retrieves only the regions visible from the target viewpoint, while tiled PRoPE establishes explicit geometric correspondence between retrieved memories and target-view attention. To model dynamic worlds, we further disentangle persistent static environments from evolving dynamic entities, storing stable scene structures in the patch memory while representing movable objects and characters using lightweight multi-view \textbf{reference tokens}. Finally, we develop a \textbf{progressive long-horizon distillation framework}, building on Perceptual Flow Matching~\cite{zhao2026perceptual} and recent causal video-distillation pipelines~\cite{zhu2026causal,zhao2026causalforcingplusplus}, that transforms a high-quality bidirectional world model into an efficient few-step fully causal generator, enabling stable real-time interaction without sacrificing long-range consistency.

Extensive experiments demonstrate that, when trained on diverse data collected from Unreal simulation environments, open-world games, and internet videos, Matrix-Game 3.5 achieves strong performance in long-horizon scene recall, geometry-consistent camera control, subject-consistent generation, prompt-driven world evolution, and stable real-time open-world interaction. These results demonstrate that explicitly unifying memory, geometry, and dynamic world representation provides an effective foundation for scalable, controllable, and persistent interactive world simulation.

Our main contributions are summarized as follows:

\begin{itemize}

\item We propose a geometry-aware persistent world representation that unifies long-horizon memory and geometry-consistent camera control; its patch-memory and tiled-PRoPE components introduce no additional learnable parameters on top of the backbone. Built on patch memory and tiled PRoPE, it enables explicit 3D memory retrieval and geometry-aware attention across viewpoint changes.

\item We propose a static-dynamic disentangled world modeling strategy with lightweight multi-view reference tokens for dynamic subject representation. Combined with motion-aware filtering and leakage-resistant subject training, it effectively suppresses ghosting and identity drift during long-horizon interactive generation.

\item We develop a progressive long-horizon real-time distillation framework that combines Perceptual Flow Matching with curriculum self-rollout DMD. It transfers a bidirectional world model into a few-step fully causal interactive generator, enabling minute-long real-time generation with long-horizon consistency and accurate camera control.
\end{itemize}

\section{Method}

Matrix-Game 3.5 consists of three tightly coupled components. First, we introduce a pose-aware sequence representation that unifies heterogeneous inputs into a single self-attention sequence with shared spatiotemporal and camera-aware coordinates. Second, we develop a unified memory system that combines geometry-aware patch memory, context memory, and reference-token memory to preserve both scene geometry and dynamic-subject identity over long-horizon interactions. Finally, we present a progressive distillation framework that converts a bidirectional video diffusion model into a few-step fully causal autoregressive generator for real-time interactive generation.

\subsection{Pose-aware Sequence Representation}

Matrix-Game 3.5 arranges all token types --- the anchor frame, the clean context frames, the retrieved memory patches, and the noisy target frames --- into a single pose-aware token sequence. Because tiled PRoPE, our camera conditioning introduced below, tiles the camera projection over the very channels that already carry the spatiotemporal RoPE, every token receives both a camera pose and a RoPE spatiotemporal coordinate within this one sequence. We first describe how the camera pose is injected into attention, and then how these coordinates are assigned across the sequence.

\subsubsection{Camera-aware Attention}

\begin{figure}[t!]
    \centering
    \includegraphics[width=\linewidth]{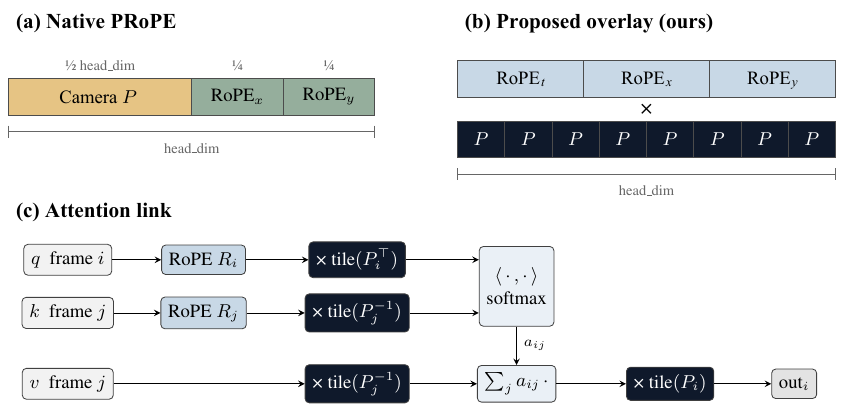}
    \caption{\textbf{Injection of camera
    projection geometry into spatiotemporal RoPE.}
    (a) Native PRoPE, designed for unordered multi-view images in novel-view
    synthesis where no time axis is needed, splits the head dimension into disjoint
    blocks: camera projection takes half, and two quarter blocks encode the 2D patch
    row and column; camera and position occupy disjoint subspaces and never multiply.
    (b) \emph{Tiled PRoPE}, our design for video generation, tiles the camera projection
    over \emph{all} channels and multiplies it onto the full spatiotemporal RoPE (frame axis
    included), allocating no dedicated subspace, so the pretrained RoPE layout stays
    intact. (c) One attention link between
    frame $i$ (query) and frame $j$ (key/value): $q$ and $k$ are first RoPE-rotated,
    then multiplied by $\mathrm{tile}(P_i^\top)$ and $\mathrm{tile}(P_j^{-1})$; a
    single softmax thus carries both the relative time $\Delta t=i-j$ and the
    relative pose $M_{ij}=P_iP_j^{-1}$. Values are transported through the same
    relative pose ($v\!\leftarrow\!P^{-1}$, $\mathrm{out}\!\leftarrow\!P$), with no
    learnable parameters.}
    \label{fig:prope-injection}
\end{figure}

Matrix-Game 3.5 first requires explicit camera control over the generated video. Moreover, since the tokens in this sequence come from different cameras and different times, we want the interaction between frames to be pose-aware rather than to rely on the token grid of the native 3D RoPE~\cite{rope} alone. We build on PRoPE~\cite{li2025prope}, an existing scheme that turns camera geometry into a relative position code: every latent frame is associated with a full world-to-image projection matrix $P=\mathrm{lift}(K)\cdot T^{wc}$, where $K$ is the $3\times3$ intrinsics matrix expressed in pixel coordinates, $T^{wc}=(T^{cw})^{-1}$ is the world-to-camera extrinsics --- the inverse of the camera-to-world pose $T^{cw}$ produced by our annotation pipeline and used by the memory modules --- and $\mathrm{lift}(\cdot)$ pads $K$ to an invertible $4\times4$ matrix with a unit bottom-right entry and zeros elsewhere, so that $P$ maps homogeneous world coordinates to homogeneous image coordinates and carries no projective scale ambiguity; the attention score between two frames then carries their relative projection $M=P_iP_j^{-1}$, jointly encoding relative rotation, translation, and intrinsics (the exact matrix conventions are given in Appendix~\ref{app:rope-intact}). Native PRoPE, designed for novel-view synthesis (NVS), splits the attention head dimension into disjoint blocks --- half for the camera projection and two quarters for the 2D row and column positions --- so camera and position never multiply and there is no time axis (Figure~\ref{fig:prope-injection}a).
This native design suits NVS, where the inputs are unordered views, but it cannot be transplanted directly to video generation, for two reasons. Carving out half of the head dimension for the camera would take those channels away from the backbone's pretrained 3D spatiotemporal RoPE, whose frame axis encodes the temporal ordering that video generation depends on; and a disjoint split, such as that adopted by DreamX~\cite{dreamx2026world}, leaves no way for a single token to carry a camera pose and a spatiotemporal position at once, as required by a sequence that mixes memory, anchor, and target tokens. We therefore introduce \emph{tiled PRoPE}, which leaves the RoPE layout intact and instead tiles the camera projection across all head channels, multiplying it onto the full spatiotemporal RoPE including its frame axis, without carving out a separate subspace (Figure~\ref{fig:prope-injection}b). Concretely, after the standard spatiotemporal RoPE rotation the queries, keys, and values are further multiplied by the camera matrices --- queries by $P^{\top}$, keys and values by $P^{-1}$, with the attention output mapped back by $P$ --- so a single softmax carries both the relative time $\Delta t=i-j$ and the relative pose $M$, adding no learnable parameters (Figure~\ref{fig:prope-injection}c). Appendix~\ref{app:rope-intact} verifies that the temporal structure of the original RoPE is preserved under this overlay.

To address numerical instability caused by large camera transformations, Matrix-Game 3.5 recenters all camera poses with respect to the first target frame of the current generation window and applies a direction-preserving logarithmic compression only to the metric-scale translation component $\mathbf{t}$:
\begin{equation}
\tilde{\mathbf{t}}
=
\frac{\log\!\left(1+\lVert\mathbf{t}\rVert\right)}{4\,\lVert\mathbf{t}\rVert}\,\mathbf{t},
\label{eq:log-compression}
\end{equation}
which maps the translation magnitude to $\log(1+\lVert\mathbf{t}\rVert)/4$ while keeping the translation direction unchanged. This preserves the ordering and direction of small camera motions while strongly compressing large translations, preventing distant historical viewpoints from introducing numerical instability.

\subsubsection{Pose-aware Sequence Layout}

\begin{figure}[ht!]
    \centering
    \includegraphics[width=\linewidth]{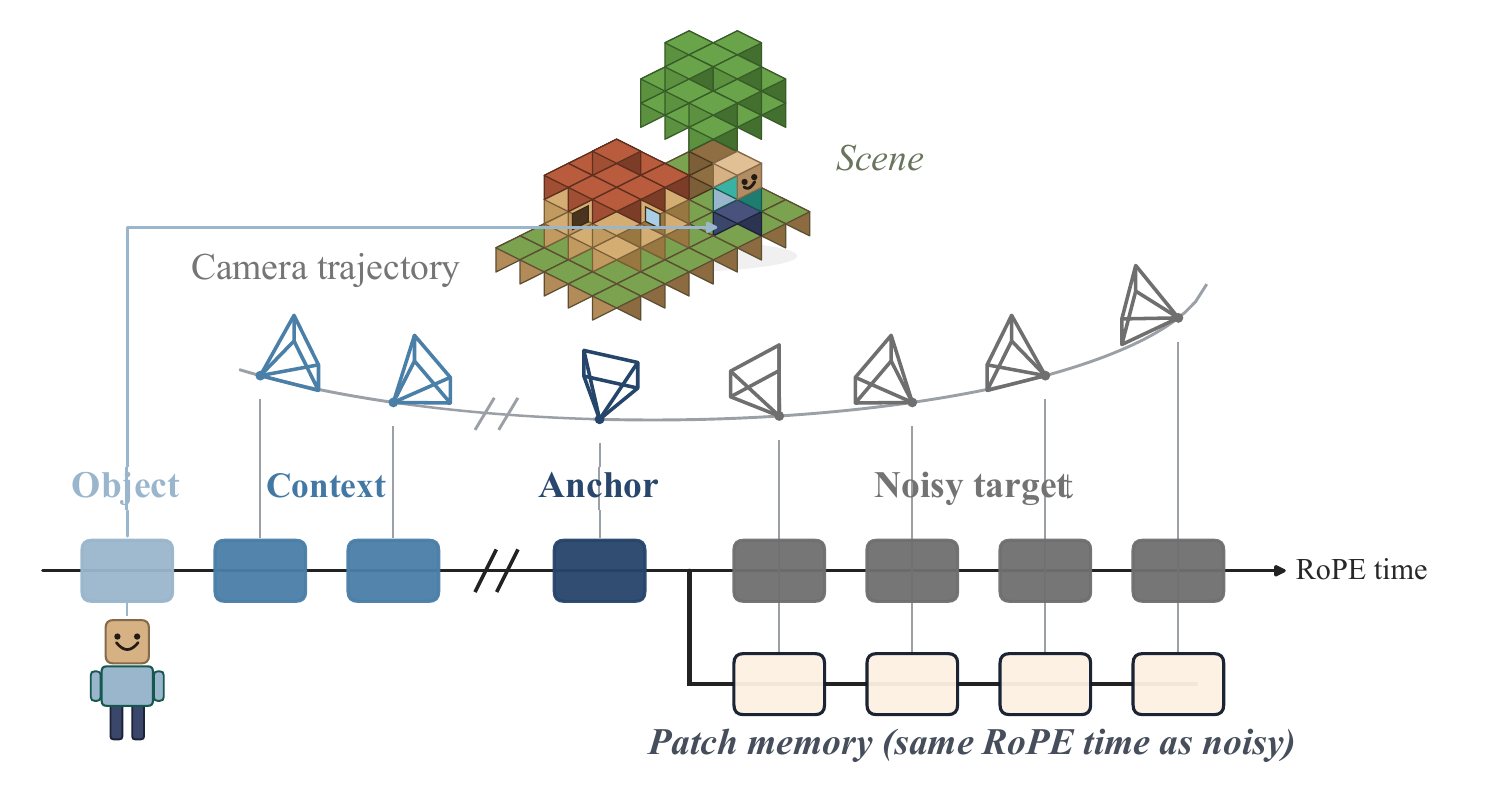}
    \caption{\textbf{The pose-aware token sequence.} All token types share
    one sequence, each carrying its own RoPE time and camera pose. \emph{Top:} the cameras
    that produce the sequence move along one trajectory
    around the scene, and the context frames (blue), the anchor (purple), and the noisy
    target frames (gray) each observe the scene from their own pose; the reference-token
    block (teal) is a prefix for the dynamic subject, shown here linked to the character
    in the scene; its tokens
    inherit the camera pose of the current anchor rather than observing the scene from a pose
    of their own. \emph{Bottom:} every camera maps
    straight down to its position on the shared RoPE timeline. The context frames keep
    their true past RoPE times, separated from the generation window by a time jump
    (\texttt{//}); the anchor---the most recent latent frame---keeps its true RoPE time,
    immediately preceding the generation window. The noisy target frames and the
    patch-memory tokens reuse the \emph{same} target camera poses at the \emph{same} RoPE
    time.}
    \label{fig:sequence-layout}
\end{figure}

At each step of long-horizon autoregressive generation, the model must reconcile several heterogeneous sources at once: faithful unwarped observations of the past, the most recent clean latent frame for local continuity, a persistent representation of the dynamic subject, geometry-aligned memory of previously seen scene content, and the noisy frames currently being denoised. Rather than routing these through separate cross-attention branches, Matrix-Game 3.5 places all of them into the single pose-aware sequence introduced above and lets one self-attention stack relate them directly through their shared RoPE coordinates and relative camera poses; the text prompt, injected through the backbone's native cross-attention, is the only condition outside this sequence. In this design, each conditioning source is incorporated simply by inserting its tokens with appropriately assigned coordinates and camera poses, and the memory mechanisms are characterized precisely by their assignment rules in Section~\ref{sec:memory}.

Concretely, the layout follows a small set of rules. The context frames and the anchor keep their true historical RoPE timestamps and relative camera poses. The noisy target frames occupy the RoPE times of the generation window, each with its target camera pose. Patch-memory tokens reuse the RoPE time and camera pose of the target frame they support, placed at the floating-point spatial coordinates where their content projects into that view. The reference tokens sit at fixed negative RoPE times before the video timeline and inherit the anchor's camera pose. These rules give every token a place in one pose-aware sequence (Figure~\ref{fig:sequence-layout}), and the memory mechanisms in the next section simply populate these coordinates.

\subsection{Unified Memory System}
\label{sec:memory}

Built on the pose-aware sequence, the unified memory system combines a geometry-aware patch memory for static scene recall, a lightweight context memory for complementary historical observations, and a reference-token memory for persistent dynamic-subject identity, explicitly disentangling static geometry from dynamic content during long-horizon generation.

\subsubsection{Patch Memory}
In autoregressive long-video generation, the current segment must not only remain visually consistent with previously generated content, but also recover scene structures that were observed earlier when the camera revisits the same location. Existing approaches can largely be divided into two categories. Explicit memory methods represent scenes using points, splats, or other geometric primitives, providing strong geometric constraints but primarily targeting static scene reconstruction. Implicit memory methods instead preserve entire video frames or latent features as memory, offering greater generative flexibility, yet relying on the network itself to learn cross-view correspondences. We observe that an intermediate memory granularity—patch-level memory—has received relatively little attention. Rather than treating an entire frame as an indivisible memory unit or maintaining a complete global 3D representation, we follow MosaicMem~\cite{yu2026mosaicmem} and store localized image patches as the fundamental unit of memory, allowing them to be selectively retrieved, spatially aligned, and reused when needed.

\begin{figure}[t!]
    \centering
    \includegraphics[width=\linewidth]{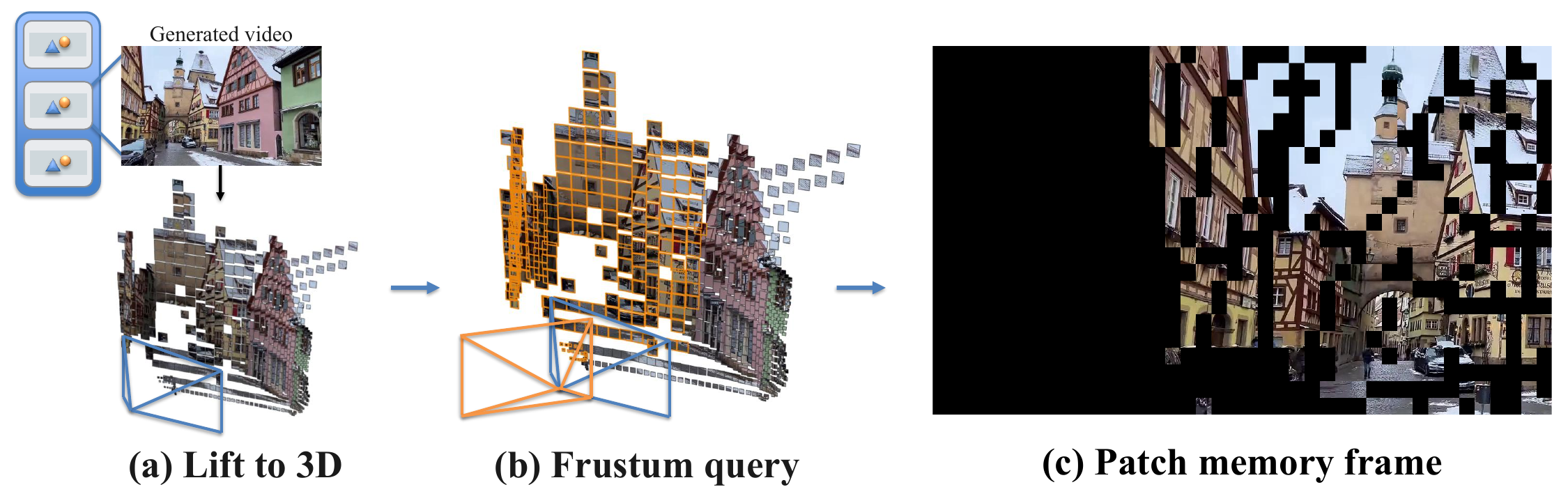}
    \caption{\textbf{The patch-memory mechanism.} (a) \emph{Lift to 3D}:
    historical latent patches are back-projected into a common 3D space using
    metric depth together with camera intrinsics and poses. (b) \emph{Frustum
    query}: the target camera frustum queries this patch cloud, and a z-buffer
    under the target view keeps, for every target latent location, only the
    surface closest to the camera while discarding occluded candidates. (c)
    \emph{Patch memory frame}: the selected patches are scattered into an aligned
    memory canvas under the target viewpoint; regions that are occluded or have
    never been observed are left empty (black) and dropped from the memory token
    sequence; the diffusion model synthesizes them from the current latent,
    text prompt, and surrounding context.}
    \label{fig:mosaic-mechanism}
\end{figure}

Our adaptation of this patch-level memory offers a principled combination of the complementary strengths of explicit and implicit memory, and further integrates it into a pose-aware autoregressive sequence for streaming world generation. Its first stage resembles an explicit-memory pipeline: historical patches are lifted into a common 3D space using metric depth together with camera intrinsics and poses. Its second stage resembles an implicit-memory pipeline: the retrieved and geometrically aligned patches are injected into the diffusion transformer as reference conditions rather than hard reconstruction targets. This design enables the model to exploit accurate geometric correspondence for reliable scene recall while retaining the flexibility to synthesize novel content, dynamic objects, and unseen regions according to the text prompt. Moreover, because memory is represented at the patch level rather than as complete frames or globally reconstructed geometry, the memory naturally supports sparse retrieval, localized editing and manipulation, and robust long-horizon memory updates without accumulating global reconstruction errors over repeated revisits.

Concretely, the patch memory addresses a simple but fundamental question: for every latent patch in the target frame, where should its memory be retrieved from within the previously observed video? Instead of learning this correspondence implicitly, we compute it directly using camera geometry. Every historical frame is associated with its camera-to-world pose, camera intrinsics, and a metric depth map. Given a latent patch from a historical frame, we first recover its center in image space according to the VAE downsampling ratio, then back-project it into the historical camera coordinate system using the corresponding depth and camera intrinsics. The resulting 3D point is transformed into world coordinates through the historical camera pose, and finally projected into the target camera using the target pose and intrinsics. This geometric lifting-and-reprojection procedure establishes an explicit patch-wise correspondence between historical observations and the target viewpoint (Figure~\ref{fig:mosaic-mechanism}). Throughout, the stored unit is the VAE latent patch: each memory entry keeps the latent feature of one cell of the VAE latent grid --- the same representation consumed by the diffusion transformer, so retrieval requires no re-encoding. Memory entries are treated as fixed conditioning, detached from the computation graph.

During memory fusion, multiple historical patches may project onto the same target latent location. To resolve occlusions and conflicting observations, we perform a z-buffer selection under the target viewpoint, retaining only the surface closest to the target camera while discarding farther candidates. The selected latent feature is written into an aligned memory canvas under the target view, while a reverse mapping records both the source frame and the original patch location from which the memory originates. Regions that are occluded, geometrically unreliable, or have never been observed leave holes on the memory canvas; these holes are dropped from the token sequence rather than kept as empty placeholder tokens, so the memory track avoids doubling the sequence length, and the diffusion model synthesizes the corresponding regions conditioned on the current noisy latent, text prompt, and surrounding context. Consequently, memory is reused only where reliable geometric evidence exists, while newly emerging content remains fully generative.

Upstream of this reprojection, the pipeline selects a small set of complementary historical frames for each target latent frame. Simply choosing the nearest camera poses often results in highly redundant viewpoints with poor overall coverage. Instead, we adopt a coverage-aware selection strategy that progressively chooses the historical frames contributing the largest additional coverage over the target latent grid. Formally, given the already-selected set $\mathcal{S}$, the next frame is chosen greedily as
\begin{equation}
f_{k}
=
\arg\max_{f\in\mathcal{H}\setminus\mathcal{S}}
\Big|\,C(f)\setminus\bigcup_{s\in\mathcal{S}}C(s)\Big|,
\label{eq:coverage-selection}
\end{equation}
where $\mathcal{H}$ denotes the set of stored historical frames and $C(f)$ the set of target latent cells that receive a valid, unoccluded depth reprojection from frame $f$; the selection is performed independently for every target latent frame and stops when the marginal coverage gain vanishes or the per-target budget of five frames (Section~\ref{sec:experiments}) is reached. The selected frames are then processed through geometric reprojection, z-buffer fusion, and latent scattering to construct two outputs: an aligned memory canvas under the target viewpoint and a patch-wise correspondence map recording the origin of every retrieved memory token. The aligned memory canvas is provided to the diffusion transformer as spatial conditioning, while the correspondence map is used to modify the positional encoding of memory tokens, enabling the model to reason about their geometric relationship with the current latent tokens. Combined with dynamic-object masking and metric-scale depth alignment, and guarded by the visibility consistency checks above --- target-view z-buffer occlusion testing together with the discarding of invalid, out-of-frame, or geometrically unreliable projections --- this mechanism provides a geometrically reliable, locally controllable, and robust patch-level spatial memory for long-horizon autoregressive video generation. Within the pose-aware sequence, each retrieved patch declares its position by geometry rather than by storage order: it takes the RoPE timestamp of the target frame it supports rather than its original historical time, together with the floating-point spatial coordinate at which its source content lands in the target view. For valid projections, these fractional coordinates index a higher-resolution RoPE frequency table for sub-grid precision, similar to PE-Field~\cite{bai2025pefield}; when sub-grid coordinates are unavailable but the projected target latent cell remains valid, the token falls back to the nearest native integer-grid coordinate, while geometrically unreliable projections are discarded entirely, as described above. A patch is thus addressed by where it should appear in the frame being generated rather than by where it was stored.

\subsubsection{Context Memory}
The patch memory supplies target-aligned local content, but that content is warped into the current view; to also give the model faithful, unwarped observations of the scene, the model additionally conditions on a small set of clean historical latent frames placed before the anchor. Context memory consists of exactly this small set of clean historical latent frames, hereafter referred to as context frames; we use these two terms consistently throughout the paper. The anchor is the most recent clean latent context immediately preceding the generation window; it contains a single latent frame or, during streaming rollout, a short multi-frame latent block (see Section~\ref{sec:experiments}). Context frames are drawn from the earlier history of the same clip. Rather than the nearest frames, which tend to be redundant, context frames are selected so that their camera poses cover the viewpoints of the window being generated, providing complementary past observations of the scene. Within the pose-aware sequence, each context frame keeps its own coordinates: it retains its true historical RoPE timestamp --- so the relative temporal gaps between the older context, the anchor, and the generating window are preserved rather than collapsed to consecutive indices --- together with its relative camera pose. The number of context frames is kept limited, far smaller than the number of frames being generated, making context memory a lightweight complement to the patch memory: it contributes whole, unwarped observations at their original time and viewpoint, while the patch memory supplies dense content warped into the current view.

\subsubsection{Static-Dynamic Disentangled Memory}

\begin{figure}[ht!]
    \centering
    \includegraphics[width=\linewidth]{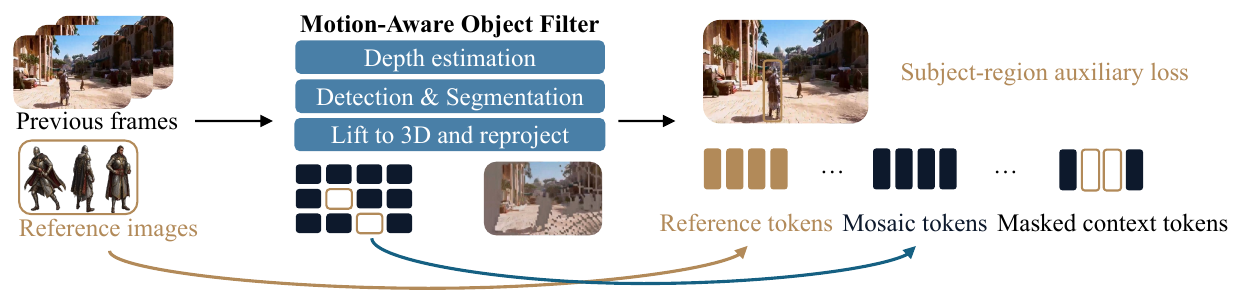}
    \caption{
    Overview of the static-dynamic disentangled memory system. A motion-aware
    object filter separates static scene observations from dynamic subjects.
    Static regions are fused into the patch memory, while dynamic subjects are
    represented by reference-token memory. Subject-region masks are also
    used for context masking and an auxiliary training loss.
    }
    \label{fig:decoupled-memory}
\end{figure}

Long-horizon video generation requires a memory system that can preserve both
scene-level stability and object-level temporal consistency. These two
requirements correspond to different temporal structures. Static scene content,
including background geometry, global layout, and stationary objects, should
remain spatially consistent across time. Dynamic objects, in contrast, preserve
their identity and appearance while undergoing time-dependent changes in
position, pose, and visibility.

The memory system separates these two types of temporal structure
(Figure~\ref{fig:decoupled-memory}). The static
branch is built on the patch memory, a geometry-aware mechanism for retrieving and
reusing previous observations of the stable scene. The dynamic branch is
represented by sequence-level reference tokens, which provide persistent identity
and appearance cues for dynamic subjects.

Without such a separation, dynamic objects can be accumulated into a static
spatial memory. The same object observed at different time steps may then appear
at different locations or with different poses. Memory fusion can conflate these
temporal states with independent static observations, leading to duplicated
objects, ghosting artifacts, incorrect occlusions, and ambiguous supervision.

\paragraph{Motion-aware Object Filter}

The motion-aware dynamic object filter serves as the interface between the
static and dynamic memory branches. It applies YOLO segmentation and tracking~\cite{sapkota2025ultralytics} to
common movable categories, including humans, vehicles, and animals, and
estimates whether each tracked object is static or dynamic within a local video
block. Regions classified as static are retained in the patch-memory branch,
whereas regions classified as dynamic are masked before memory fusion.

The filter is based on motion rather than semantic category. A stationary
foreground object remains a useful spatial reference, whereas an object from the
same category should be excluded from static memory when it exhibits motion.
Motion is therefore used as the criterion for static-memory validity.

For each track, the filter uses the predicted depth map together with camera
intrinsics and extrinsics to test geometric consistency across nearby frames. Let
$M_i$ denote the segmentation mask in frame $i$, $D_i$ the corresponding depth
map, and $(K_i, E_i)$ the camera intrinsics and extrinsics. Mask samples are
lifted from frame $i$ to 3D and reprojected to frame $j$, yielding a directional
overlap score
\begin{equation}
s_{i \rightarrow j}
=
\frac{
\left|\Pi_j\!\left(\mathcal{U}_i(M_i, D_i, K_i, E_i)\right) \cap M_j\right|
}{
\left|\Pi_j\!\left(\mathcal{U}_i(M_i, D_i, K_i, E_i)\right)\right|
},
\end{equation}
where $\mathcal{U}_i(\cdot)$ denotes depth-based unprojection and $\Pi_j(\cdot)$
denotes projection into frame $j$. Before computing the overlap, we discard
points with invalid depth, points whose depth in the target camera is
non-positive, and projections falling outside the target image; tracks with
fewer than $N_{\min}$ valid projected samples are marked as uncertain. The
symmetric static-consistency score is computed as
\begin{equation}
s_{ij} = \min \left(s_{i \rightarrow j}, s_{j \rightarrow i}\right),
\end{equation}
so that a high $s_{ij}$ indicates cross-frame geometric agreement, i.e., a
static observation, rather than motion. Tracks with sufficient temporal support and consistently high $s_{ij}$ are
retained as static observations, while tracks with low $s_{ij}$ are excluded
from the patch memory. Uncertain tracks are kept outside the static branch. The same
filtering criterion is used in training and inference. At inference time, the
filter is applied only when the frames of a completed rollout segment are
registered into the patch memory --- once per segment rather than per generated
frame --- so its cost is amortized outside the per-frame denoising loop and does
not enter the real-time budget of Section~\ref{sec:realtime}.

\paragraph{Reference Token Memory}

The dynamic branch of the memory system is implemented with sequence-level
reference tokens. In the current third-person game setting, the dynamic subject
refers to the controllable character whose identity and appearance should remain
consistent throughout the rollout. Reference tokens serve as persistent
representations of this subject, providing long-term identity and appearance
memory even when it changes position, pose, or visibility.

Four reference images are encoded through the VAE and mapped to a compact latent
token budget. At inference time, the four reference views are extracted only
from the causally available initialization and history, and remain fixed
throughout the rollout; no future frames are ever accessed. Each reference image uses roughly one quarter of a single frame's
token budget, so the four-reference prefix is comparable in length to one
additional context frame.
Each reference patch token receives additive reference-index, type, and local spatial embeddings; these embeddings are the only learnable parameters the framework adds on top of the backbone. The reference views are further assigned fixed-gap temporal RoPE coordinates: the $r$-th reference is placed at $t_r=-r\Delta_{\mathrm{ref}}$, while the original video timeline remains unchanged. This gap-timing RoPE distinguishes the reference views and keeps their temporal placement independent of the absolute rollout step. 
For tiled PRoPE, all reference tokens inherit the camera transform of the current clean anchor rather than a fixed identity camera. Since the clean anchor advances with each autoregressive window, the relative camera transform between the reference prefix and the target frames remains local to the current generation window, instead of growing with the accumulated camera motion over the full rollout.

During both training and inference, detected subject regions in the anchor and
context frames are replaced with background patches sampled from the same scene
rather than with zeros or blurred noise, so that the conditioning statistics
match between the two phases and direct appearance leakage from the visible
conditioning frames is prevented. During training, we additionally apply
stronger random mask dilation and perturbation, forcing the model to recover
subject appearance from the reference tokens rather than from residual context
pixels. A
subject-region auxiliary diffusion loss is computed on the subject mask to
increase the gradient weight on the dynamic subject and encourage consistent
subject-level modeling:
\begin{equation}
\mathcal{L}
=
\mathcal{L}_{\mathrm{full}}
+
\lambda_{\mathrm{sub}}
\frac{
\sum_{p} M_{\mathrm{sub}}(p)
\left\| \mathbf{v}_{\theta}(p) - \mathbf{v}(p) \right\|_2^2
}{
\sum_{p} M_{\mathrm{sub}}(p) + \varepsilon
}.
\end{equation}
Here $M_{\mathrm{sub}}$ denotes the subject-region mask, $\mathbf{v}_{\theta}$
and $\mathbf{v}$ are the backbone's native flow-matching velocity prediction
and its regression target --- the same parameterization used by the base model
and by the distillation stages in
Section~\ref{sec:progressive_distillation} --- $\mathcal{L}_{\mathrm{full}}$ is
the standard flow-matching loss over all tokens, and
$\lambda_{\mathrm{sub}}$ controls the additional subject-region weight.

The prefix-token design avoids an additional cross-attention module. A
cross-attention formulation would introduce a separate alignment problem between
reference space and video space, while also adding a parameterized conditioning
path that can be bypassed through shortcut solutions. The prefix formulation
keeps the reference branch inside the original transformer token space and
allows it to interact with the generated video through the same self-attention
stack.

\subsection{Progressive Distillation}
\label{sec:progressive_distillation}

\begin{figure}[ht!]
    \centering
    \includegraphics[width=0.7\linewidth]{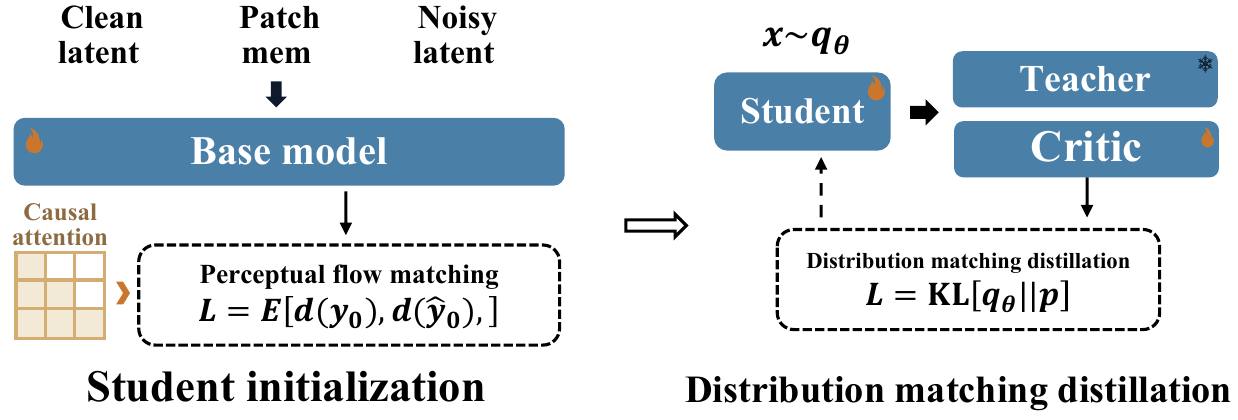}
    \caption{
    \textbf{Overview of our two-stage progressive distillation pipeline.}
    First, causal adaptation uses teacher-forced perceptual flow matching to
    obtain a high-quality few-step causal initializer. Second, self-rollout DMD
    directly optimizes inference-time autoregression through a curriculum that
    progressively distills classifier-free guidance, camera control, and
    memory-conditioned generation.
    }
    \label{fig:distillation_pipe}
\end{figure}

We use the two-stage pipeline in Figure~\ref{fig:distillation_pipe}. A
teacher-forced causal-adaptation stage first provides a high-quality few-step
initializer, after which self-rollout distribution matching directly optimizes
the inference-time autoregressive process.

\paragraph{Causal Adaptation}
We reformulate generation as chunk-wise causal denoising. For target chunk
$i$, the teacher-forced condition is
\begin{equation}
    \mathcal{H}_{i}^{\mathrm{gt}}
    =
    \left(
    \mathbf{x}_{<i}^{\mathrm{gt}},
    \mathbf{p}_{\leq i},
    \mathbf{m}_{i}^{\mathrm{gt}},
    \mathbf{r}_{i}^{\mathrm{gt}},
    \mathbf{c}
    \right),
\end{equation}
where $\mathbf{p}_{\leq i}$ denotes causal camera control,
$\mathbf{m}_{i}^{\mathrm{gt}}$ is the query-aligned patch memory,
$\mathbf{r}_{i}^{\mathrm{gt}}$ is the chunk-local context frame --- selected by the same coverage rule as in Section~\ref{sec:memory}, at one frame per chunk --- and
$\mathbf{c}$ is the text condition. Neither attention nor memory retrieval can
access future chunks. Given noise $\boldsymbol{\epsilon}$, we construct
\begin{equation}
    \mathbf{x}_t^i
    =
    (1-t)\mathbf{x}_i+t\boldsymbol{\epsilon}.
\end{equation}
Following perceptual flow matching (PFM)~\cite{zhao2026perceptual}, we recover
the clean prediction
\begin{equation}
    \hat{\mathbf{x}}_{0,\theta}^{i}
    =
    \mathbf{x}_t^i
    -t\,\mathbf{v}_{\theta}
    \left(\mathbf{x}_t^i,t\mid\mathcal{H}_{i}^{\mathrm{gt}}\right),
\end{equation}
and constrain flow matching in a pretrained perceptual feature space rather
than regressing velocity only in the VAE latent space. Let $\mathcal{D}$ be the
frozen VAE decoder, $\Phi_{\ell}$ the $\ell$-th feature block of a frozen
perceptual model, and
$\hat{\mathbf{x}}_{0,\theta}^{\leq i}=
[\mathbf{x}_{<i}^{\mathrm{gt}},\hat{\mathbf{x}}_{0,\theta}^{i}]$ the complete
teacher-forced causal chain. We optimize
\begin{equation}
    \mathcal{L}_{\mathrm{PFM}}
    =
    \mathbb{E}_{\mathbf{x},i,t,\boldsymbol{\epsilon}}
    \left[
    \frac{1}{|\mathcal{S}|}
    \sum_{\ell\in\mathcal{S}}
    d\!\left(
    \Phi_{\ell}\!\left(\mathcal{D}
    (\hat{\mathbf{x}}_{0,\theta}^{\leq i})\right),
    \Phi_{\ell}\!\left(\mathcal{D}
    (\mathbf{x}_{\leq i}^{\mathrm{gt}})\right)
    \right)
    \right],
    \label{eq:perceptual_causal_adaptation}
\end{equation}
where $d$ measures feature distance and $\mathcal{S}$ contains the selected
perceptual blocks. This single objective simultaneously learns causal
denoising and few-step generation, yielding an efficient, high-quality causal
initializer.

\paragraph{Self-Rollout Distribution Matching}
The second stage removes teacher forcing and performs DMD~\cite{yin2024one,yin2024improved}
on the student's own autoregressive trajectory. For chunk $i$, the online
condition becomes
\begin{equation}
    \widehat{\mathcal{H}}_{i}^{\theta}
    =
    \left(
    \hat{\mathbf{x}}_{<i}^{\theta},
    \mathbf{p}_{\leq i},
    \hat{\mathbf{m}}_{i}^{\theta},
    \hat{\mathbf{r}}_{i}^{\theta},
    \mathbf{c}
    \right),
\end{equation}
where both patch memory and context frames are retrieved online from the
causally visible generated history. The objective matches the student
distribution $q_{\theta,t}$ to the bidirectional teacher distribution
$p_{\phi,t}$ under this shared condition:
\begin{equation}
    \mathcal{L}_{\mathrm{DMD}}
    =
    \mathbb{E}_{i,t}
    \left[
    w(t)\,
    \mathrm{KL}
    \left(
    q_{\theta,t}
    \left(\cdot\mid\widehat{\mathcal{H}}_{i}^{\theta}\right)
    \,\Vert\,
    p_{\phi,t}
    \left(\cdot\mid\widehat{\mathcal{H}}_{i}^{\theta}\right)
    \right)
    \right].
    \label{eq:world_model_dmd}
\end{equation}
Applying this objective to autoregressive video requires care because the
student and scorers maintain different memory states. The student updates online
memory from generated chunks, while feeding the same potentially drifted history
to the bidirectional scorers would compromise the supervision. We therefore
share only stable external conditions---the initial memory, anchor frame, text
prompt, and camera trajectory---while allowing the student to update its memory
and keeping scorer memory fixed. This provides reliable scores without forcing
incompatible internal trajectories to match.

To stabilize training under this interface, we adopt a condition curriculum:
first distilling classifier-free guidance and camera control without online
memory, and then gradually introducing patch memory and context frames.
Following HiAR~\cite{zou2026hiar}, we keep the autoregressive prefix and
chunk-local context at the next noise level during each denoising substep while
retaining a clean anchor. This represents imperfect generated history with
appropriate uncertainty, reducing long-horizon drift and encouraging the use of
geometrically aligned patch memory.

Together, perceptual causal adaptation and curriculum-based self-rollout DMD
yield a three-step causal generator that preserves quality and control over
minute-long, real-time interactive streams.

\section{Data Infrastructure}
\label{sec:data}
Memory-aware interactive world models require supervision beyond raw videos. In particular, our model relies on three complementary forms of supervision that are unavailable in large-scale video corpora: geometry supervision for long-horizon memory retrieval and geometry-consistent camera control, semantic supervision for temporally coherent world generation, and identity supervision for persistent dynamic-subject modeling. To this end, we build an offline data infrastructure that augments raw videos with geometry, language, and object-level annotations, followed by quality-aware data curation. The resulting training corpus provides unified supervision for the patch memory, tiled PRoPE, and the lightweight multi-view reference tokens.

\subsection{Geometry Annotation for Geometry-Aware Generation}

\begin{figure}[t]
    \centering
    \begin{minipage}[c]{0.40\linewidth}
        \centering
        {\scriptsize
        \begin{tikzpicture}[x=1cm, y=1cm, >=stealth, font=\scriptsize,
            box/.style={draw=black!75, rounded corners=2pt, align=center,
                        inner sep=3pt, minimum height=0.42cm}]
            \node[font=\scriptsize\itshape, anchor=west] at (0,0.52) {Video frames};
            \draw[fill=gray!12, draw=black!60] (0,0) rectangle (5.0,0.30);
            \foreach \x in {0.25,0.50,...,4.80} \draw[black!35] (\x,0) -- (\x,0.30);
            \draw[fill=blue!14, draw=black!60, rounded corners=1pt]
                (0,-0.60) rectangle (2.9,-0.24);
            \node at (1.35,-0.42) {Segment $k$};
            \draw[fill=blue!14, draw=black!60, rounded corners=1pt]
                (2.1,-1.04) rectangle (5.0,-0.68);
            \node at (3.6,-0.86) {Segment $k$+1};
            \fill[orange!50, opacity=0.30] (2.1,-1.04) rectangle (2.9,0.30);
            \draw[black!50, dashed] (2.1,-1.04) rectangle (2.9,0.30);
            \node[font=\tiny\itshape, text=orange!55!black, rotate=90] at (2.55,-0.37) {Overlap};
            \node[box, fill=violet!14] (vggt) at (1.2,-1.98) {VGGT-Omega};
            \node[box, fill=violet!14] (da3)  at (3.8,-1.98) {DA3};
            \draw[->] (2.5,-1.04) -- (2.5,-1.48);
            \draw[black!75] (1.2,-1.48) -- (3.8,-1.48);
            \draw[->] (1.2,-1.48) -- (vggt.north);
            \draw[->] (3.8,-1.48) -- (da3.north);
            \node[font=\tiny\itshape, anchor=west] at (2.62,-1.24) {Each segment};
            \node[box, fill=gray!7] (vout) at (1.2,-2.80) {Scale-free\\pose + depth};
            \node[box, fill=gray!7] (dout) at (3.8,-2.80) {Metric depth};
            \draw[->] (vggt.south) -- (vout.north);
            \draw[->] (da3.south)  -- (dout.north);
            \node[box, fill=orange!18, minimum width=4.6cm] (mrg) at (2.5,-3.60) {Rescale to metric};
            \node[box, fill=teal!16,   minimum width=4.6cm] (sim) at (2.5,-4.24) {Sim(3) alignment};
            \node[box, fill=gray!22,   minimum width=4.6cm] (out) at (2.5,-4.88) {Depth, pose, intrinsics, ...};
            \draw[->] (vout.south) -- (vout.south |- mrg.north);
            \draw[->] (dout.south) -- (dout.south |- mrg.north)
                node[midway, right, font=\tiny\itshape] {Scale factor};
            \draw[->] (mrg.south) -- (sim.north);
            \draw[->] (sim.south) -- (out.north);
        \end{tikzpicture}}
    \end{minipage}\hfill
    \begin{minipage}[c]{0.012\linewidth}
        \centering
        \rule{0.4pt}{5.3cm}
    \end{minipage}\hfill
    \begin{minipage}[c]{0.575\linewidth}
        \centering
        \includegraphics[width=\linewidth]{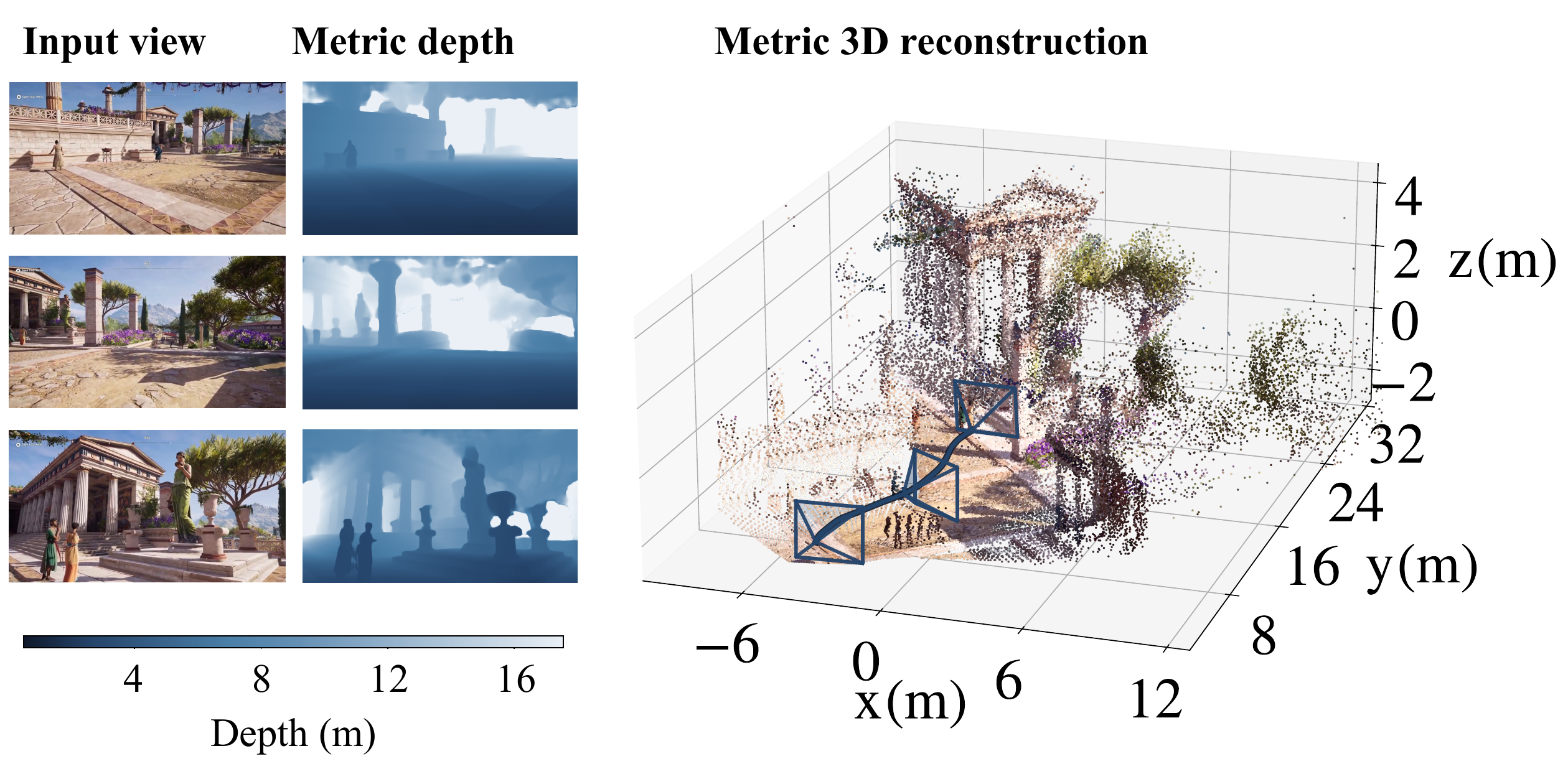}
    \end{minipage}
    \caption{\textbf{Metric-scale geometric
    annotation.} Left: the annotation pipeline --- chunkwise VGGT-Omega inference
    over overlapping segments, per-segment metric anchoring with the Depth
    Anything~3 metric branch, and Sim(3) cross-segment alignment with a single
    global metric scale. Right: a reconstruction example --- three views of the
    same scene, each with its input frame and per-frame estimated metric
    depth, and the three views
    back-projected into a shared metric frame (axes in estimated meters), yielding a
    colored point cloud, a smoothed camera trajectory, and the
    camera frusta of the three views. The pipeline outputs metric depth,
    camera-to-world poses, camera intrinsics, and scale and quality
    diagnostics.}
    \label{fig:metric-annotation}
\end{figure}
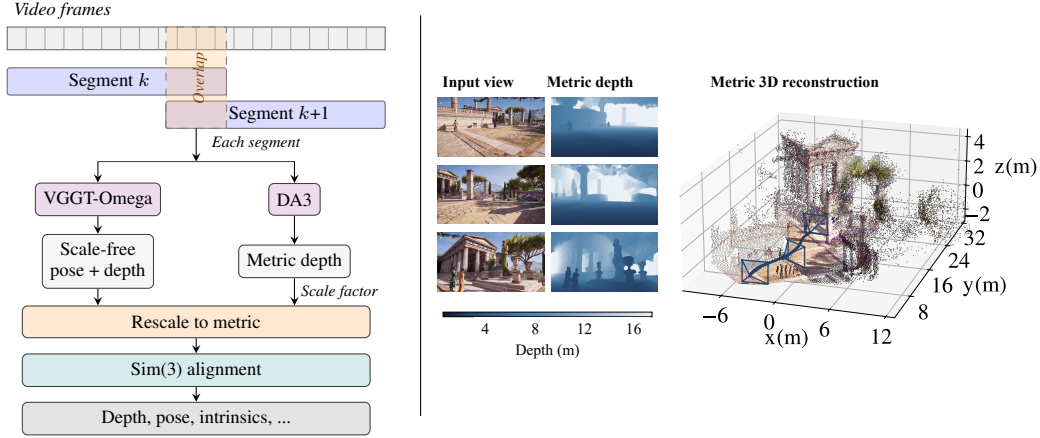

The patch memory and tiled PRoPE require accurate geometric supervision to establish consistent 3D correspondences across long-horizon generation. Since raw videos do not provide camera poses, metric depth, or camera intrinsics, we build an offline geometry annotation pipeline based on VGGT-Omega~\cite{wang2026vggtomega}, summarized in Figure~\ref{fig:metric-annotation}. Long videos are processed in overlapping temporal chunks to avoid memory limitations and accumulated reconstruction errors.

Because VGGT-Omega predicts geometry only up to an unknown scale, each chunk is first rescaled to metric units using the metric branch of Depth Anything~3~\cite{lin2025depthanything3} as a scale anchor. Neighboring chunks are then stitched through Sim(3) alignment on their overlapping frames; this pairwise alignment allows each chunk its own corrective scale so that the stitched trajectory remains continuous. We then robustly aggregate the per-chunk metric-scale estimates into a single global scale and apply it exactly once, uniformly rescaling the aligned camera translations and depth maps together, which balances inter-chunk continuity with overall metric consistency without reintroducing per-chunk scale drift. The final annotation consists of camera-to-world poses $T^{cw}$, camera intrinsics, and metric depth maps, providing estimated metric-scale geometric supervision for geometry-aware memory retrieval and camera control.

\subsection{Semantic Annotation for World Generation}

Interactive world generation requires temporally coherent semantic supervision describing both persistent scene layouts and dynamic object evolution. Rather than generating independent frame-level captions, we annotate each temporal window using Gemma 4 26B A4B IT (\texttt{google/gemma-4-26b-a4b-it})~\cite{team2026gemma}: a sparse set of frames sampled within the window is provided to the model as a multi-image input, and one prompt is generated per window, matching the temporal granularity of the downstream generator.

Each prompt jointly describes the static environment together with the identities, motions, trajectories, pose changes, and interactions of all dynamic objects. Degenerate descriptions such as ``no motion'' are explicitly suppressed, while subjective descriptions unrelated to scene dynamics are removed. The generated window-level prompts are expanded to all frames within the corresponding window and aligned with the geometric annotations, providing temporally stable semantic supervision for world generation. Appendix~\ref{app:prompt-examples} shows annotated windows from game and real-world sources.

\subsection{Identity Annotation for Dynamic Subject Modeling}

To support persistent dynamic-subject modeling, we construct an offline identity annotation pipeline that extracts object detections, segmentation masks, track identities, and multi-view reference images from each video. An offline YOLO-based pipeline~\cite{sapkota2025ultralytics} first produces temporally consistent object tracks for common dynamic categories. For third-person game videos, the controllable character is automatically identified according to tracking quality, temporal duration, spatial prominence, image-center proximity, and mask quality, eliminating the need for manual identity annotation.

Multi-view reference images are then extracted from diverse timestamps to maximize viewpoint and appearance diversity. During training, DINO features~\cite{simeoni2025dinov3} are used to preferentially sample mutually dissimilar references, providing complementary identity and appearance cues for training the lightweight multi-view reference tokens.

\subsection{Scene-Quality Data Curation}

Large-scale automatic annotation inevitably introduces inaccurate geometry and low-quality samples. We therefore perform quality-aware data curation before training. Each annotated clip is scored along four complementary perspectives: reconstruction quality (reconstruction confidence, trajectory smoothness, intrinsics stability), camera behavior (viewpoint revisits and a sweet-spot reward on camera motion), visual quality (image quality, depth validity), and scene suitability (person count, sky fraction, near-field occlusion).

Two further terms correct biases of the scoring itself: video length prevents short fragments, which accumulate less error, from scoring spuriously high on the other axes, and camera exploration rewards the spatial extent of the trajectory, counteracting a systematic preference for nearly static cameras. Curation acts as a conjunction of per-metric gates rather than a single weighted score: each axis is scored with hard per-metric criteria, and a clip is retained in the training corpus only if it passes all of them.
\section{Experiments}
\label{sec:experiments}

\subsection{Implementation Details}
\noindent\textbf{Camera- and Memory-augmented Base Model.}
We build Matrix-Game 3.5 upon Wan2.2-TI2V-5B~\cite{wan2025wan} and jointly optimize camera conditioning and the patch memory in a single training stage. Camera intrinsics and extrinsics are injected in every DiT block through tiled PRoPE, while geometrically retrieved historical observations are fused into query-aligned memory latents and processed by the same self-attention stack. We fine-tune the full DiT while keeping the text encoder and VAE frozen. Training is conducted at a resolution of $1280\times704$ with 21 target latent frames per sample. Each sample contains one anchor latent frame and up to five historical context frames selected by trajectory coverage. With probability 0.8, the anchor jointly encodes a short multi-frame latent block, mirroring streaming rollout during inference, where each generation window's anchor is the last latent frame of the previous window; otherwise it encodes a single frame, mirroring the first window initialized from one image. For each target latent, we retrieve up to five historical candidates and fuse them into target-aligned memory using depth-aware z-buffering. For a training sample with 21 target latent frames, the assembled sequence thus comprises the 21 noisy target frames, one anchor, up to five context frames, a reference prefix of roughly one frame's worth of tokens, and the retained memory tokens; because unobserved holes are dropped, the memory track adds at most one token per covered target cell, so the total sequence length is bounded by roughly $2.3\times$ that of the target track alone even at full coverage, and is shorter in practice as coverage is partial. The patch memory is enabled for 80\% of training iterations and omitted for the remaining 20\%, while tiled PRoPE camera conditioning is retained in both cases. Dynamic-object regions are filtered from the patch memory, and text conditioning is dropped with probability 0.1. We use AdamW with a constant learning rate of $5\times10^{-5}$, weight decay of 0.01, and gradient clipping at 0.5. Training uses BF16 precision, DeepSpeed ZeRO-2, and gradient checkpointing.

\noindent\textbf{Distillation Model.}
For perceptual causal adaptation, training is conducted at
$1280\times704$. Each causal training window spans 22 positions on the latent timeline: one clean anchor frame followed by seven autoregressive chunks of three target latent frames each.
Chunk $i$ can attend to the anchor and preceding ground-truth chunks, whereas future chunks are masked. We query patch memory for every target latent, and one context frame is selected for each three-latent chunk. The memory
and context tokens are drawn only from causally visible history and are exposed only to their corresponding target latent or chunk. We convert the predicted flow to clean latent frames using the exact scheduler noise level, decode the complete causal chain with the frozen VAE, and average feature distances over all blocks of a frozen InternVideo2-1B encoder~\cite{wang2024internvideo2}.
This stage is trained for 10,000 optimization steps using AdamW with a learning rate of $5\times10^{-6}$.

For self-rollout DMD, the student generates three latent frames per chunk with three denoising steps. At every substep, the rolling autoregressive prefix and selected context frame are noised to the next scheduler level, while the first anchor frame remains clean. On memory-conditioned iterations, each generated chunk is decoded and published to the online patch memory and context memory before the next chunk is queried. We begin DMD with both memory conditions disabled and progressively
increase the probability of enabling them together. The real scorer and student rollout both use CFG with guidance scale 3. At each student denoising step, we evaluate conditional and unconditional branches with a shared negative prompt, form $\mathbf{v}^{-}+3(\mathbf{v}^{+}-\mathbf{v}^{-})$, and retain gradients only through the conditional branch. We train self-rollout DMD for 1,000 outer optimization steps, using learning rates of $2\times10^{-6}$ and $4\times10^{-7}$ for the student and fake scorer, respectively. The fake scorer is updated at every outer step, whereas the student generator is updated once every five outer steps, yielding a $5{:}1$ fake-scorer-to-student update ratio.

\subsection{Quantitative Results}
\noindent\textbf{Evaluation Protocol and Metrics.}
Following the one-minute world-model benchmark introduced by SANA-WM~\cite{zhu2026sana}, we evaluate minute-long generation on the Simple- and Hard-Trajectory splits. For pose accuracy, the estimated and target camera trajectories are aligned by a similarity transform before measuring rotation error (R, mean angular deviation in degrees), translation error (T, mean Euclidean distance between aligned camera centers in meters), and camera-motion consistency error (CMC, mean Frobenius norm of the difference between the full camera pose matrices); lower values indicate more faithful camera control. VBench Overall~\cite{huang2024vbench} aggregates visual quality, temporal consistency, and motion-related dimensions, with higher values being better. Efficiency is reported as peak memory and video throughput, with each method run on the number of GPUs listed in Table~\ref{tab:merged_quantitative}. Revisit consistency compares generated frame pairs captured from nearly identical camera poses: following the SANA-WM protocol, same-pose pairs are selected as temporally distant frames of the same rollout whose aligned camera poses nearly coincide. PSNR and SSIM~\cite{wang2004ssim} reward pixel and structural agreement, respectively, whereas LPIPS~\cite{zhang2018lpips} measures perceptual distance. Finally, temporal degradation compares VBench imaging quality in the first and last 10-second windows, with $\Delta\mathrm{IQ}=\mathrm{IQ}_{0\text{--}10}- \mathrm{IQ}_{50\text{--}60}$; a smaller drop indicates better long-horizon stability.

\definecolor{rankoneblue}{HTML}{8FCBFF}
\definecolor{ranktwoblue}{HTML}{BFE0FF}
\definecolor{rankthreeblue}{HTML}{E3F2FF}

\newcommand{\bestcell}[1]{\cellcolor{rankoneblue}#1}
\newcommand{\secondcell}[1]{\cellcolor{ranktwoblue}#1}
\newcommand{\thirdcell}[1]{\cellcolor{rankthreeblue}#1}

\begin{table*}[t]
\centering
\caption{Quantitative comparison on the one-minute benchmark of SANA-WM~\cite{zhu2026sana}. Bold Res marks 720p, and \#G is the number of GPUs used for inference. Matrix-Game 3.5 is evaluated before distillation; see Section~\ref{sec:realtime} for the distilled model. Pose Acc. reports R in degrees, plus T/CMC; VBench reports Overall score only. Revisit consistency is measured by PSNR/SSIM/LPIPS on same-pose pairs, and temporal degradation compares imaging quality between the first and last 10-second windows. Mem/Tput are peak GB and videos/hour, measured on a node with 8 H100s. Blue highlights mark top-three entries, with darker blue indicating a better rank; tied entries share a rank.}
\label{tab:merged_quantitative}
\small
\setlength{\tabcolsep}{3.0pt}
\renewcommand{\arraystretch}{1.05}
\resizebox{\textwidth}{!}{
\begin{tabular}{@{}lccccccccccccccc@{}}
\toprule
& & & & \multicolumn{3}{c}{Pose Acc. ($\downarrow$)} & \multicolumn{1}{c}{VBench ($\uparrow$)} & \multicolumn{2}{c}{Efficiency} & \multicolumn{3}{c}{Revisit consistency} & \multicolumn{3}{c}{Temporal degradation} \\
\cmidrule(lr){5-7} \cmidrule(lr){8-8} \cmidrule(lr){9-10} \cmidrule(lr){11-13} \cmidrule(lr){14-16}
Method & Param & Res & \#G & R & T & CMC & Overall & Mem$\downarrow$ & Tput$\uparrow$ & PSNR$\uparrow$ & SSIM$\uparrow$ & LPIPS$\downarrow$ & IQ$_{0-10}\uparrow$ & IQ$_{50-60}\uparrow$ & $\Delta$IQ$\downarrow$ \\
\midrule
\multicolumn{16}{@{}l}{\emph{Simple-Trajectory Split}} \\
\midrule
Infinite-World~\cite{gao2026infiniteworlds} & 1.3B & 480p & 1 & 16.55 & 1.98 & 2.08 & 79.18 & \secondcell{53.5} & 5.9 & 12.60 & 0.284 & 0.595 & \bestcell{73.93} & 67.22 & 6.72 \\
LingBot-World~\cite{team2026advancing} & 14B+14B & 480p & 8 & 10.47 & 2.01 & 2.05 & \bestcell{81.82} & 454.1 & 0.6 & \bestcell{14.59} & \secondcell{0.366} & \bestcell{0.394} & \thirdcell{73.46} & \bestcell{73.42} & \bestcell{0.04} \\
HY-WorldPlay~\cite{sun2025worldplay} & 8B & 480p & 8 & 17.89 & 2.36 & 2.45 & 68.82 & 215.5 & 1.1 & 12.83 & 0.321 & 0.616 & 70.08 & 46.50 & 23.59 \\
\midrule
Matrix-Game 3.0~\cite{wang2026matrixgame3} & 5B & \textbf{720p} & 8 & 12.96 & 1.83 & 1.92 & 78.53 & 106.2 & 3.1 & 12.29 & 0.326 & 0.553 & 69.07 & 66.66 & \thirdcell{2.41} \\
SANA-WM~\cite{zhu2026sana} & 2.6B & \textbf{720p} & 1 & \thirdcell{7.59} & \thirdcell{1.59} & \thirdcell{1.63} & 79.29 & \bestcell{51.1} & \bestcell{24.1} & 14.16 & \thirdcell{0.333} & \thirdcell{0.458} & 72.63 & 68.84 & 3.79 \\
SANA-WM + refiner~\cite{zhu2026sana} & 2.6B+17B & \textbf{720p} & 1 & \secondcell{4.50} & \secondcell{1.39} & \secondcell{1.41} & \secondcell{80.62} & \thirdcell{74.7} & \secondcell{22.0} & \thirdcell{14.46} & 0.292 & 0.479 & 73.37 & \secondcell{72.21} & \secondcell{1.17} \\
\textbf{Matrix-Game 3.5} & 5B & \textbf{720p} & 1 & \bestcell{1.63} & \bestcell{1.10} & \bestcell{1.11} & \thirdcell{80.14} & 77.0 & \thirdcell{10.9} & \secondcell{14.56} & \bestcell{0.439} & \secondcell{0.439} & \secondcell{73.80} & \thirdcell{70.56} & 3.24 \\
\midrule
\multicolumn{16}{@{}l}{\emph{Hard-Trajectory Split}} \\
\midrule
Infinite-World~\cite{gao2026infiniteworlds} & 1.3B & 480p & 1 & 41.31 & 2.49 & 2.84 & 79.51 & \secondcell{53.5} & 5.9 & 12.04 & 0.248 & 0.617 & \bestcell{73.79} & 69.63 & 4.16 \\
LingBot-World~\cite{team2026advancing} & 14B+14B & 480p & 8 & 18.99 & \thirdcell{1.65} & 1.81 & \bestcell{81.89} & 454.1 & 0.6 & 14.08 & \secondcell{0.332} & \bestcell{0.436} & \secondcell{73.66} & \bestcell{73.09} & \thirdcell{0.58} \\
HY-WorldPlay~\cite{sun2025worldplay} & 8B & 480p & 8 & 35.46 & 2.34 & 2.64 & 70.46 & 215.5 & 1.1 & 13.72 & \thirdcell{0.328} & 0.654 & 70.21 & 44.33 & 25.88 \\
\midrule
Matrix-Game 3.0~\cite{wang2026matrixgame3} & 5B & \textbf{720p} & 8 & 18.79 & 1.67 & 1.82 & 78.79 & 106.2 & 3.1 & 12.17 & 0.317 & 0.556 & 69.24 & 68.92 & \secondcell{0.32} \\
SANA-WM~\cite{zhu2026sana} & 2.6B & \textbf{720p} & 1 & \thirdcell{10.02} & 1.66 & \thirdcell{1.72} & 79.60 & \bestcell{51.1} & \bestcell{24.1} & \thirdcell{14.10} & 0.327 & \thirdcell{0.469} & 72.58 & 69.49 & 3.09 \\
SANA-WM + refiner~\cite{zhu2026sana} & 2.6B+17B & \textbf{720p} & 1 & \secondcell{8.34} & \secondcell{1.39} & \secondcell{1.44} & \bestcell{81.89} & \thirdcell{74.7} & \secondcell{22.0} & \bestcell{14.80} & 0.312 & \secondcell{0.458} & 73.34 & \secondcell{73.03} & \bestcell{0.31} \\
\textbf{Matrix-Game 3.5} & 5B & \textbf{720p} & 1 & \bestcell{2.70} & \bestcell{1.25} & \bestcell{1.33} & \thirdcell{80.85} & 77.0 & \thirdcell{10.9} & \secondcell{14.14} & \bestcell{0.414} & 0.474 & \thirdcell{73.52} & \thirdcell{71.12} & 2.40 \\
\bottomrule
\end{tabular}}
\end{table*}

\noindent\textbf{Comparison with State of the Art.}
As shown in Table~\ref{tab:merged_quantitative}, Matrix-Game 3.5 achieves the best camera accuracy among the compared methods on every pose metric and on both trajectory splits of the SANA-WM one-minute benchmark. On the Simple split, it obtains R/T/CMC errors of 1.63/1.10/1.11, reducing the rotation error from the next-best 4.50 to 1.63. The advantage becomes even more pronounced on the Hard split, where the rotation error is 2.70 compared with 8.34 for the runner-up, while T and CMC remain the best at 1.25 and 1.33.
These results show that the generated camera follows the requested trajectory stably even under challenging long-horizon motion. This gain does not come at the expense of appearance quality: our VBench Overall scores are 80.14 and 80.85 on the two splits, competitive with the strongest baselines while generating at 720p.

Matrix-Game 3.5 also achieves the highest revisit SSIM on both splits (0.439 and 0.414) with competitive PSNR (14.56 and 14.14), indicating strong structural recall by the patch memory. The remaining gaps in PSNR and LPIPS reflect the pixel-aligned nature of these metrics in a dynamic world: revisiting the same camera pose does not imply an identical image, as characters, vegetation, weather, and other background content continue to evolve, so pixel-aligned metrics penalize valid scene dynamics in addition to memory errors. The strong SSIM and pose results together provide a more faithful picture of geometric scene recall. Over one minute, the imaging-quality scores remain at 70.56 and 71.12 in the last 10-second window, with moderate drops of 3.24 and 2.40, further showing that visual quality does not exhibit severe cumulative degradation.

\subsection{Qualitative Results}
\begin{figure*}[ht!]
    \centering
    \includegraphics[width=\textwidth]{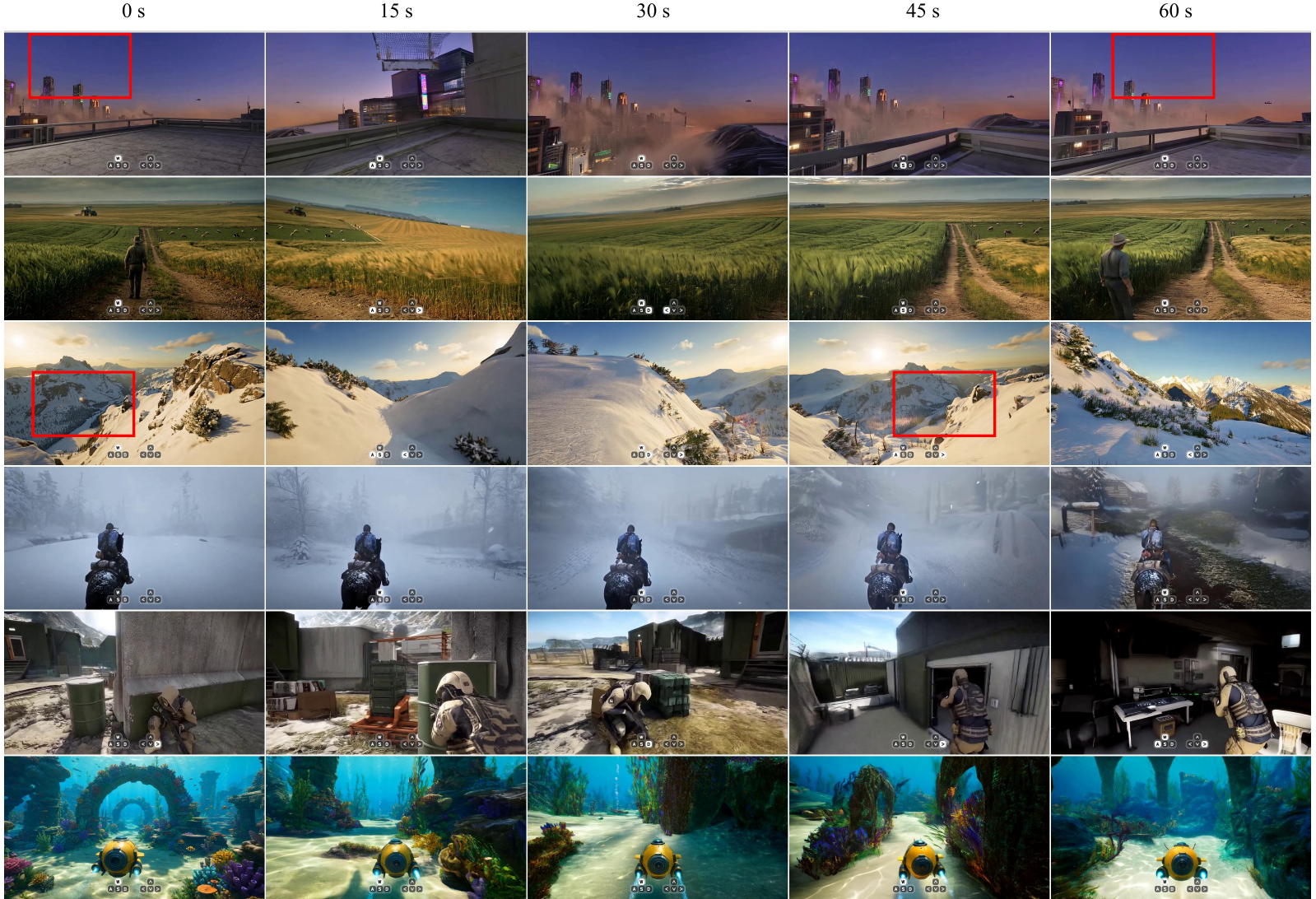}
    \caption{\textbf{Long-horizon generation in game-style environments.} Columns show frames at 0, 15, 30, 45, and 60 seconds. Scene layout and local appearance remain coherent throughout extended camera motion. Red boxes in the 60-second frames mark regions that match content observed at 0 seconds,
    highlighting long-term recall enabled by memory.}
    \label{fig:qualitative_results}
\end{figure*}

\begin{figure*}[ht!]
    \centering
    \includegraphics[width=\textwidth]{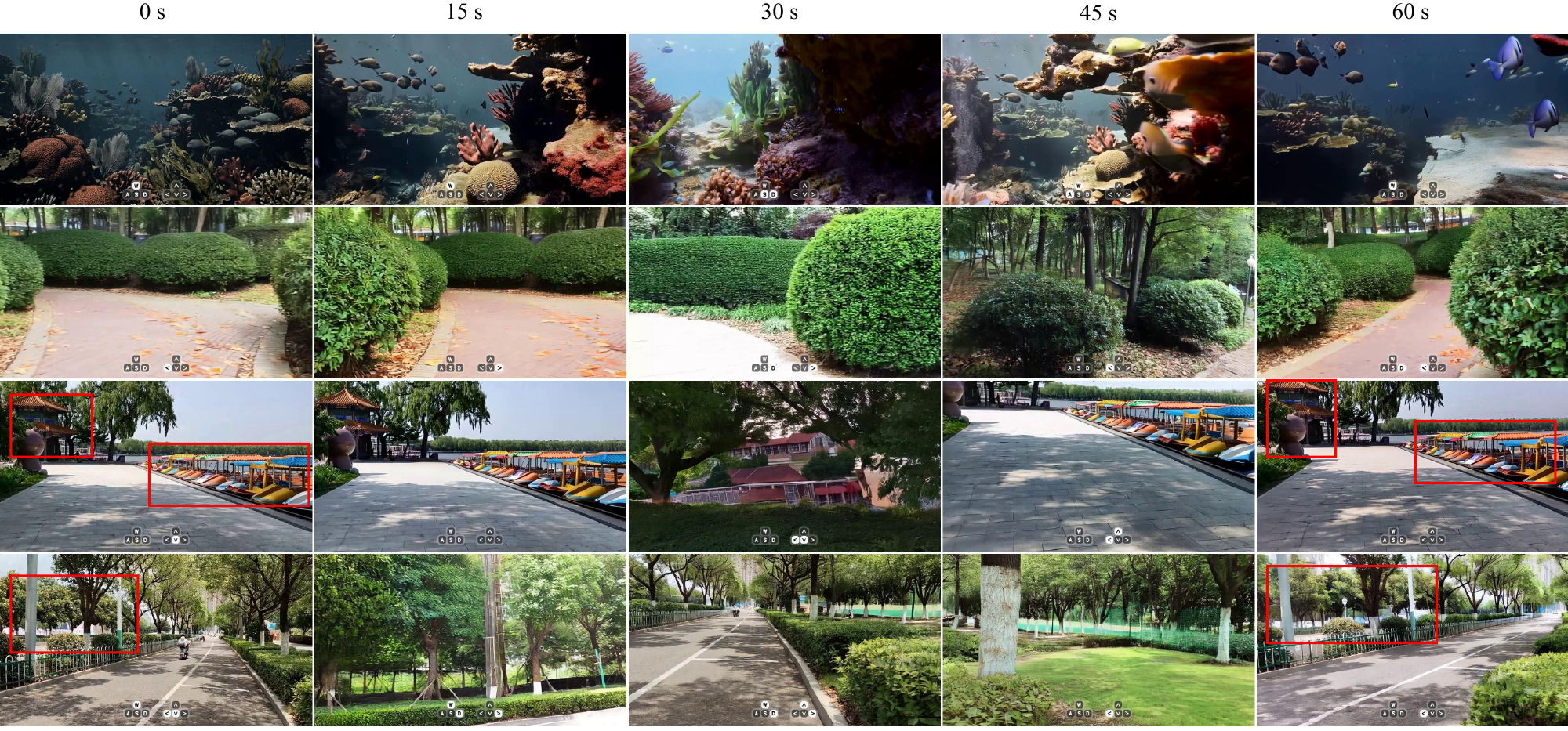}
    \caption{\textbf{Long-horizon generation in real-world environments.} Columns show frames at 0, 15, 30, 45, and 60 seconds. The model maintains a realistic appearance and stable scene geometry across sustained camera motion. Red boxes in the 60-second frames mark regions that match content observed at 0 seconds, highlighting long-term recall enabled by memory.}
    \label{fig:qualitative_results_real}
\end{figure*}

Figures~\ref{fig:qualitative_results} and \ref{fig:qualitative_results_real} present long-horizon interactive rollouts in game-style and real-world environments, respectively. Across large viewpoint changes, the patch memory preserves recognizable scene structure and local appearance, while tiled PRoPE produces smooth and stable camera motion. The late-stage frames remain sharp and coherent, and revisited content agrees closely with the initial observations, as highlighted by the red boxes. These results demonstrate that the same memory-augmented model supports persistent scene recall across both stylized game content and realistic visual domains.

\begin{figure*}[ht!]
    \centering
    \includegraphics[width=\textwidth]{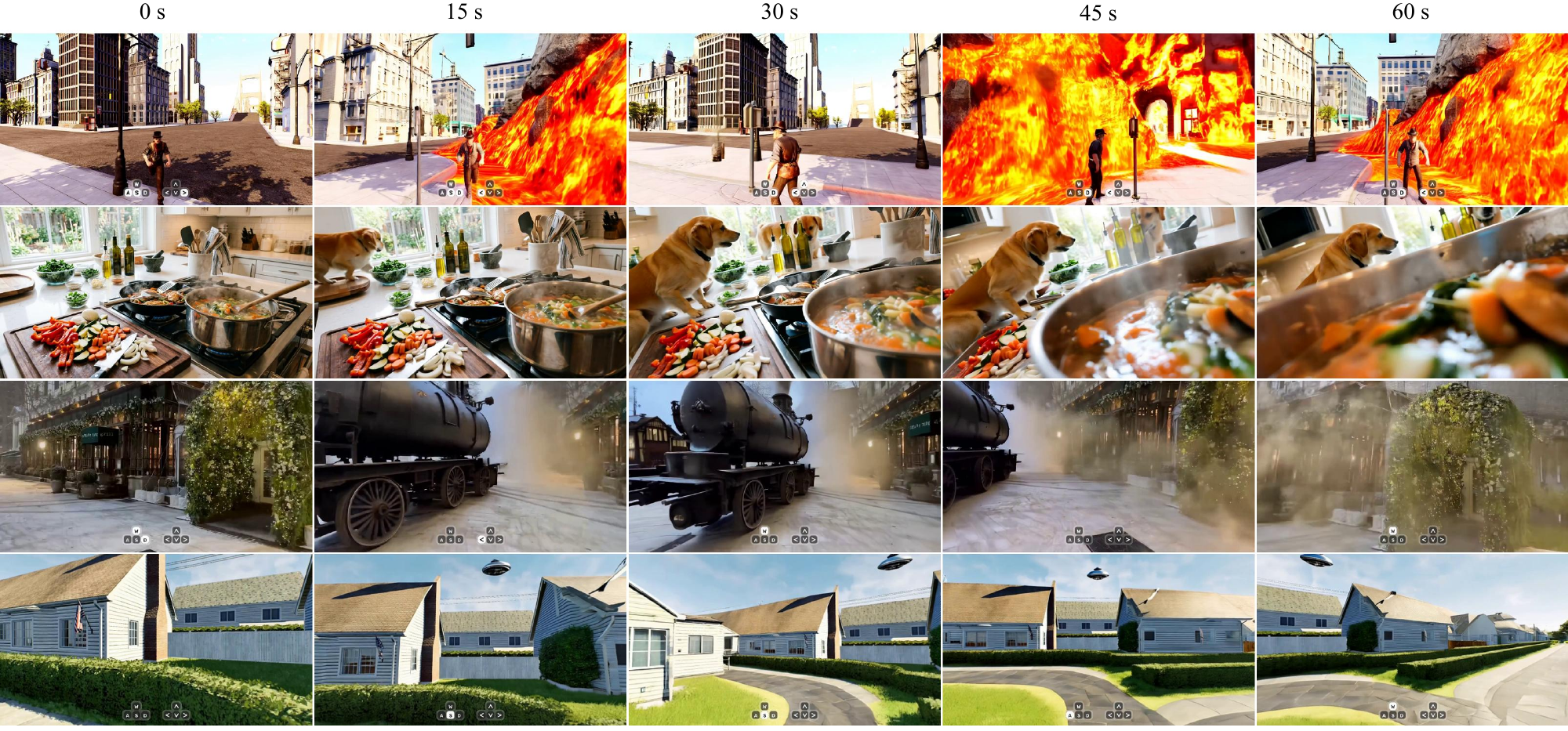}
    \caption{\textbf{Prompt-controllable interactive generation.} Columns show frames at 0, 15, 30, 45, and 60 seconds. From top to bottom, the user events are the appearance of a volcano, a dog, a train, and a UFO; each event is injected once and then carried forward by the per-segment prompt rewriting. For each rollout segment, a VLM combines the rollout context observed so far --- including pose-ordered frames retrieved from the patch memory when available --- with the camera motion and user event to rewrite a complete prompt for the next segment. The event is carried forward across segments while scene coherence and camera control are preserved.}
    \label{fig:qualitative_results_promptable}
\end{figure*}

Figure~\ref{fig:qualitative_results_promptable} further demonstrates
prompt-controllable generation. Before each autoregressive rollout segment, a VLM rewrites the complete video prompt for the next segment from the rollout context observed so far --- including pose-ordered frames retrieved by the patch memory when available --- together with the actual camera motion and a user-specified event, carrying the event forward naturally across segments.
The four rows show, from top to bottom, the appearance of a volcano, a dog, a train, and a UFO. The generated videos retain the surrounding scene and controllable camera motion while responding clearly to these events. This shows that long-term memory and camera conditioning preserve rather than suppress the pretrained model's prompt-following ability.

\begin{figure*}[ht!]
    \centering
    \includegraphics[width=\textwidth]{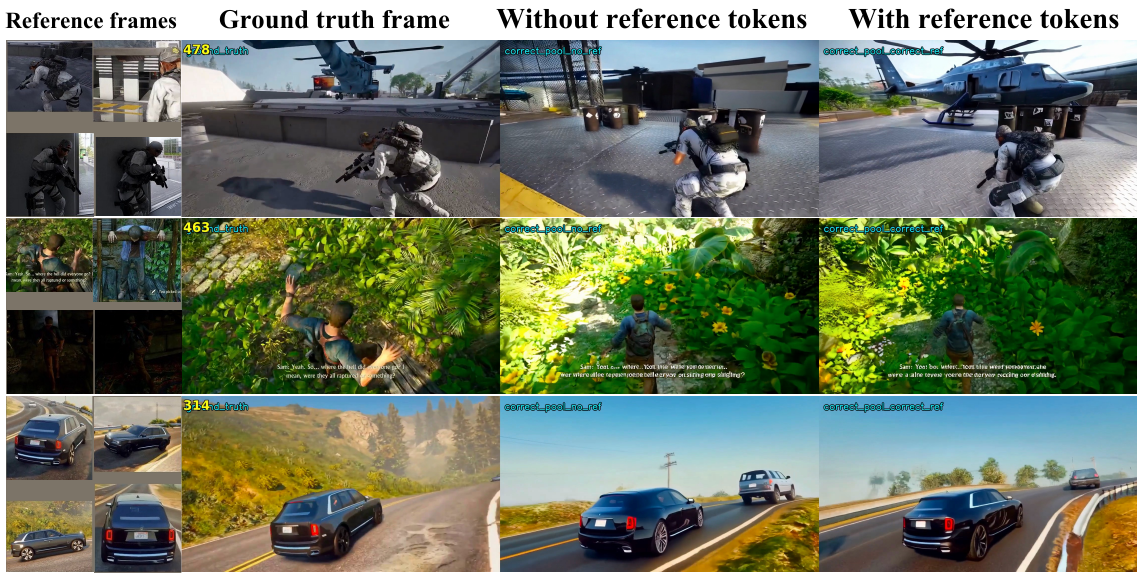}
    \caption{\textbf{Effect of dynamic-subject reference tokens.}
    We compare matched rollouts with and without the reference-token prefix while
    keeping all other conditions fixed. The reference frames (first column) are
    drawn only from the causally available initialization and history, and the
    ground-truth frame is shown only as an identity reference for visual
    comparison --- it is never provided to the model. Reference conditioning reduces identity and
    appearance drift under camera motion; the frames shown are taken 300--400 frames into the rollout. In the first, second, and third rows, respectively, it better preserves the soldier's tactical vest, the protagonist's back-mounted weapon, and the vehicle's silver side trim.}
    \label{fig:reference_token_ablation}
\end{figure*}

Figure~\ref{fig:reference_token_ablation} compares generation with and without the dynamic-subject reference prefix. We keep the initial state, camera trajectory, text prompt, and random seed fixed, and remove only the reference tokens at inference time. Without reference conditioning, the controllable subject gradually drifts in identity and local appearance under viewpoint changes and extended rollout. Enabling reference tokens preserves the subject's characteristic appearance more consistently while leaving the background evolution and camera motion unchanged. This comparison supports the role of reference tokens as a persistent appearance memory for dynamic subjects.

\begin{figure*}[ht!]
    \centering
    \includegraphics[width=\textwidth]{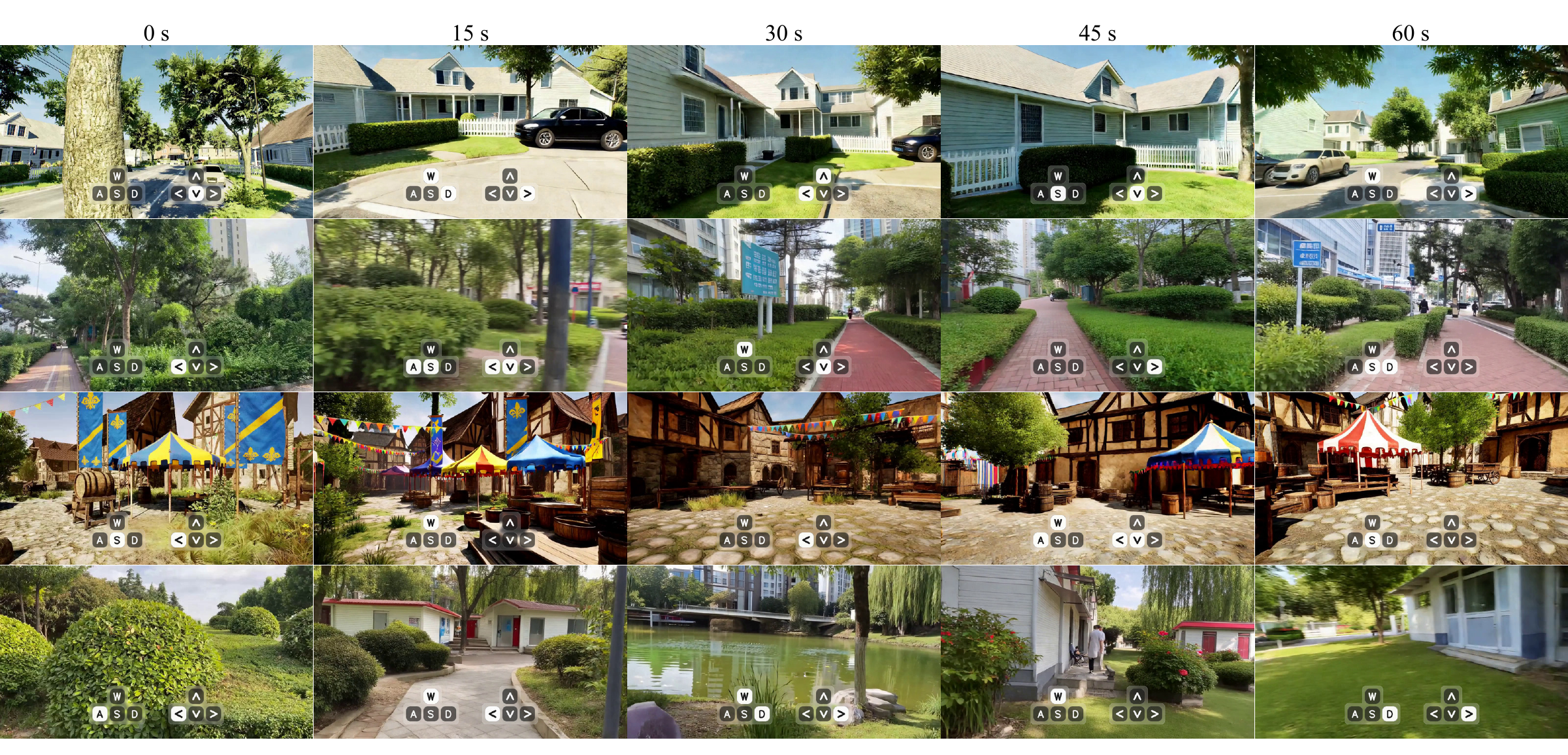}
    \caption{\textbf{Long-horizon generation with the distilled model.} Columns show frames at 0, 15, 30, 45, and 60 seconds. Rows cover four diverse outdoor environments spanning real-world and game scenes. The three-step causal student follows sustained camera motion while preserving scene structure, appearance, and fine detail throughout the rollout.}
    \label{fig:qualitative_results_distilled}
\end{figure*}

Figure~\ref{fig:qualitative_results_distilled} shows long-horizon rollouts from the final three-step distilled model. The four examples span a suburban neighborhood, an urban greenway, a medieval-style market, and a landscaped waterfront residence. Across approximately one minute of interactive camera motion, the distilled model retains sharp local detail and coherent scene geometry without abrupt scene resets. This demonstrates that our distillation pipeline preserves the base model's long-horizon controllability under aggressive sampling acceleration.

\subsection{Real-Time Inference}
\label{sec:realtime}
Following Matrix-Game 3.0~\cite{wang2026matrixgame3}, we incorporate three additional inference optimizations: DiT quantization, optimized memory retrieval, and a lightweight VAE decoder. Specifically, INT8 quantization and PyTorch compilation accelerate DiT inference with negligible impact on generation quality. For memory retrieval, we reduce the number of depth candidates evaluated for each generated chunk and migrate key retrieval operations to the GPU.
With DiT inference significantly accelerated, VAE decoding becomes the primary latency bottleneck in high-resolution streaming generation. We therefore introduce MG-LightVAE, a 75\%-pruned decoder variant that further improves end-to-end throughput while maintaining high reconstruction quality.
Overall, these optimizations enable Matrix-Game 3.5 to reach an end-to-end steady-state throughput of up to 20 output frames per second at $1280\times704$ with batch size 1 on a single NVIDIA H100 GPU, including DiT denoising with the INT8-quantized three-step student, online memory retrieval, and MG-LightVAE decoding, while the VLM prompt-rewriting module runs asynchronously outside this loop. Appendix~\ref{app:collision-avoidance} further describes an optional camera collision-avoidance component that adapts the input trajectory to already-generated scene geometry during rollout.

\section{Related Work}
\subsection{Interactive Video World Models}
Latent diffusion and Diffusion Transformer backbones provide the scalable visual priors behind modern video generators~\cite{rombach2022high,peebles2023scalable}. Systems such as Sora, HunyuanVideo, and Wan demonstrate high-fidelity generation through large-scale data and model scaling ~\cite{openai2024worldsim,kong2024hunyuanvideo,wan2025wan}, but are primarily designed for offline clip synthesis rather than closed-loop interaction. Interactive world models instead predict future observations conditioned on actions, camera motion, or high-level instructions. Genie and Genie~2 demonstrated action-controllable environment generation from video data
~\cite{bruce2024genie,parkerholder2024genie2}; GameGen-X and GameFactory focus on interactive game-video generation~\cite{che2024gamegen,yu2025gamefactory}; DeepVerse formulates 4D world modeling through autoregressive video generation
~\cite{chen2025deepverse}; and
OASIS and Matrix-Game~2.0 advanced open-source, real-time autoregressive interaction~\cite{oasis2024,he2025matrix}. More recent systems have pushed toward minute-scale and higher-resolution rollouts. WorldPlay couples keyboard-and-mouse action control with a reconstituted context memory for long-horizon streaming~\cite{sun2025worldplay},
LingBot-World targets real-time minute-long simulation~\cite{team2026advancing}, SANA-WM improves the efficiency of 720p minute-scale generation with hybrid linear attention~\cite{zhu2026sana}, and Matrix-Game~3.0 combines camera-aware memory, self-corrective training, and multi-segment DMD for real-time 720p streaming~\cite{wang2026matrixgame3}. These systems establish the feasibility of long-horizon interactive generation, but jointly maintaining accurate cross-view recall, dynamic-subject identity, prompt-driven evolution, and real-time causal rollout remains difficult.

\subsection{Long-Horizon Memory and Geometric Consistency}
Long-horizon generation is limited by finite context windows and autoregressive error accumulation, and existing memories differ chiefly in the unit they store. RELIC compresses history into latent tokens inside the KV cache, encoding actions and absolute poses so that recall stays implicit~\cite{hong2025relic}. Frame-level memories instead keep whole observations and retrieve them geometrically: Context as Memory and WorldMem select historical frames by field-of-view overlap~\cite{yu2025context,xiao2025worldmem}, VMem indexes stored views by the surfels they observed so that a target view recovers exactly the frames that saw its visible surfaces~\cite{li2025vmem}, WorldPlay reconstitutes context from past frames and reframes them in time to keep long-past observations reachable~\cite{sun2025worldplay}, and Matrix-Game~3.0 retrieves view-relevant frames based on camera pose and field-of-view overlap~\cite{wang2026matrixgame3}. A frame, however, can only be recalled as a whole and at the viewpoint where it was recorded. MosaicMem moves to patch granularity, lifting localized patches into 3D and composing them in the queried view while leaving unseen regions generative~\cite{yu2026mosaicmem}. Camera control has developed in parallel, from learned camera-conditioning modules to Projective Positional Encoding (PRoPE), which represents relative camera intrinsics and extrinsics directly in attention~\cite{he2024cameractrl,li2025prope}. Matrix-Game~3.5 builds on these directions by coupling patch-level geometric
retrieval with tiled PRoPE, and by separating static scene memory from reference-token memory for dynamic subjects.

\subsection{Dynamic-Subject Control and Static-Dynamic Memory}
Object-controllable video generation has primarily followed two complementary
directions. Trajectory-conditioned methods represent boxes, masks, or tracklets
with instance-aware spatial tokens or control features, enabling local motion
control and reducing identity swaps under occlusion~\cite{wang2024boximator,li2025trackdiffusion,zhang2026tgt}.
Reference-conditioned methods instead extract appearance from one or more
subject images and inject it through reference attention, shared token spaces,
or reconstruction objectives to preserve identity across motion and viewpoint
changes~\cite{wei2024dreamvideo,mai2025contextanyone,wei2026memento}.
In interactive world models, recent methods additionally explore training-free
concept insertion, persistent object-state tokens, explicit orchestration of
3D object trajectories, or even code-based methods~\cite{akdemir2026zero,wang2026worlddirector,xiong2026actworld,wang2026scenecode}.
These approaches improve either object motion control or long-range identity,
but generally do not decide which observations should become part of a static
geometric world and which should remain independently evolving entities. We
make this separation explicit: a motion-aware 3D consistency filter admits
only stable observations into the patch memory, whereas the controllable subject
is represented by compact multi-view reference tokens in the native,
pose-aware DiT sequence. Leakage-resistant context masking and a subject-region
objective further encourage the model to recover subject appearance from the
reference memory rather than copy it from recent frames.

\subsection{Real-Time Autoregressive Video Generation and Distillation} Streaming video generation reduces response latency by producing frames or chunks causally instead of denoising an entire clip at once. Recent systems
combine short- and long-term memory, chunk-wise autoregressive diffusion,
efficient latent diffusion and one-step online processing to support
low-latency continuous video streams
~\cite{henschel2025streamingt2v,teng2025magi,HaCohen2024LTXVideo,zhuang2025flashvsr,li2026distillalign}. These works motivate causal
computation and systems-level optimization, but a high-quality bidirectional video model must still be transferred carefully into a few-step autoregressive generator.

Distribution Matching Distillation (DMD) enables one-step or few-step sampling by matching the student's generated distribution to that of a diffusion teacher ~\cite{yin2024one}. CausVid applies distribution matching to convert a bidirectional video diffusion model into a causal autoregressive generator ~\cite{yin2025slow}. Self-Forcing further trains on autoregressive student rollouts to reduce exposure bias, and Self-Forcing++ extends this strategy to
minute-scale generation~\cite{huang2025self,cui2025self}. Self Gradient Forcing restores the missing gradient path from future losses to self-generated causal memory through a bounded two-pass replay, enabling native minute-scale video extrapolation from short training windows while improving long-horizon identity, layout, and temporal consistency
~\cite{zhuang2026self}. Causal Forcing identifies the architectural mismatch between bidirectional teachers and causal students and initializes the student from an autoregressive teacher before DMD
~\cite{zhu2026causal}. Causal Forcing++ replaces costly trajectory-based initialization with causal consistency distillation, enabling scalable frame-wise one- or two-step generation~\cite{zhao2026causalforcingplusplus}. Our distillation pipeline follows this causal-distillation line while bringing
geometry-conditioned memory into the autoregressive rollout: perceptual feature-space flow matching first adapts the causal student, and curriculum self-rollout DMD then trains the few-step model on its own long-horizon memory and context distribution.

\section{Conclusion}

We presented Matrix-Game 3.5, a real-time streaming interactive world model that advances the Matrix-Game series toward geometry-aware and long-horizon consistent world simulation. At its core is a unified geometry-aware memory framework whose patch-memory and tiled-PRoPE components introduce no additional learnable parameters: the patch memory lifts historical observations into a persistent 3D patch representation and retrieves only the regions visible from the target viewpoint, while tiled PRoPE tiles camera projection geometry over the backbone's spatiotemporal RoPE, so that long-term memory and camera control are handled by the same self-attention stack without any architectural modification. On top of this representation, a static-dynamic disentangled world modeling strategy stores stable scene structure in the static patch memory and carries movable subjects through lightweight multi-view reference tokens, suppressing ghosting and identity drift during extended rollouts. Finally, a two-stage progressive long-horizon distillation framework, combining perceptual feature-space Flow Matching with curriculum-based Self-Rollout DMD, converts the bidirectional world model into a few-step fully causal generator that sustains minute-long real-time interaction. Trained on a unified corpus spanning Unreal simulation environments, open-world games, and internet videos, Matrix-Game 3.5 achieves the best camera-control accuracy among the compared methods across all pose metrics on both the Simple- and Hard-Trajectory splits of the SANA-WM one-minute benchmark, strong revisit consistency and subject preservation, and stable prompt-driven world evolution at 720p, demonstrating that explicitly unifying memory, geometry, and dynamic world representation is an effective foundation for persistent interactive world generation.

While Matrix-Game 3.5 substantially advances long-horizon interactive generation, several directions remain open. The current design persists scene structure through the patch memory and dynamic-subject identity through reference tokens; a natural next step is to give dynamic entities persistent states of their own, so that subjects leaving the camera's view keep evolving coherently and re-emerge seamlessly when the viewpoint returns. Likewise, the geometry grounding the patch memory currently comes from an external estimator; making the generator itself geometry-aware --- jointly predicting the depth and camera pose of its own generated content --- would let geometric grounding and generation reinforce each other. Extending the framework toward richer embodied action spaces and physical interaction, and scaling to larger backbones and longer horizons, are further promising directions. We hope Matrix-Game 3.5 serves as a solid foundation for future research on persistent, fully interactive world simulation.

\clearpage
\appendix

\section{Text-Prompt Annotation Examples}\label{app:prompt-examples}

\begin{figure}[h]
    \centering
    \includegraphics[width=\linewidth]{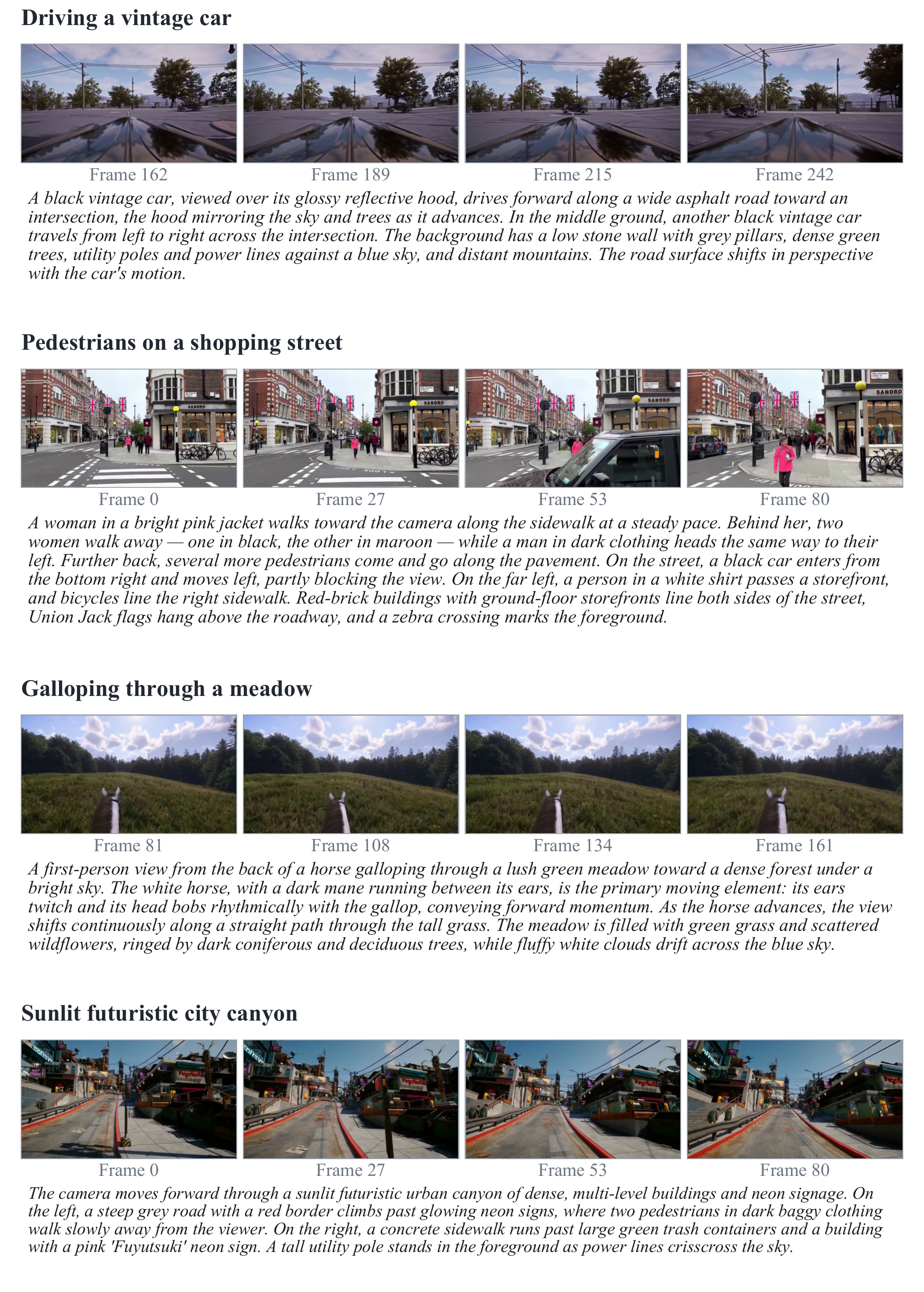}
    \caption{\textbf{Prompt-annotation examples.} Each case
    shows one annotation window as four evenly spaced frames
    (left to right) with its window-level prompt below. Across game and real-world
    sources, the prompt describes the static scene and names
    each dynamic object together with its motion.}
    \label{fig:prompt-cases}
\end{figure}

\section{Preservation of RoPE's Temporal Encoding under Tiled PRoPE}
\label{app:rope-intact}

Tiled PRoPE multiplies the camera transform onto the channels that already carry the spatiotemporal RoPE; this appendix verifies that the camera term re-weights, rather than overwrites, the temporal information those channels encode. The question arises because native PRoPE was built for novel-view synthesis, where the inputs are unordered multi-view images and the positional code carries no time axis, whereas video generation depends on the frame ordering encoded along the temporal axis of RoPE.

\paragraph{Matrix conventions.}
Attention operates on each head separately with head dimension $d_h$, which must be a multiple of 4. $\mathrm{tile}(P)\in\mathbb{R}^{d_h\times d_h}$ denotes the block-diagonal matrix that repeats the $4\times4$ projection $P$ along the diagonal, acting on each consecutive 4-channel group of a token feature as if that group were a homogeneous 4-vector: the translation column of $P$ acts on the fourth channel of each group, which plays the role of the homogeneous coordinate, so no explicit constant needs to be appended to the features. Because $\mathrm{lift}(K)$ fixes the bottom-right entry of $P$ to one, $P$ carries no projective scale ambiguity and is used without normalization. Treating per-token features as column vectors, after the spatiotemporal RoPE rotations $R_i, R_j$ the query of frame $i$ is transformed as $q\mapsto\mathrm{tile}(P_i)^{\top}R_i\,q$ and the key of frame $j$ as $k\mapsto\mathrm{tile}(P_j)^{-1}R_j\,k$, giving the attention logit
\begin{equation}
a_{ij}
=\frac{\big(\mathrm{tile}(P_i)^{\top}R_i\,q\big)^{\top}\big(\mathrm{tile}(P_j)^{-1}R_j\,k\big)}{\sqrt{d_h}}
=\frac{(R_i\,q)^{\top}\,\mathrm{tile}\!\big(P_iP_j^{-1}\big)\,(R_j\,k)}{\sqrt{d_h}},
\label{eq:tiled-prope-logit}
\end{equation}
where the second equality uses $\mathrm{tile}(A)\,\mathrm{tile}(B)=\mathrm{tile}(AB)$; each 4-channel group therefore contributes a bilinear form through the relative projection $M_{ij}=P_iP_j^{-1}$ exactly, with no transposition ambiguity between row- and column-vector conventions. Values are transformed as $v\mapsto\mathrm{tile}(P_j)^{-1}v$ and the attention output of frame $i$ is mapped back by $\mathrm{tile}(P_i)$, so values are likewise transported through $\mathrm{tile}(P_iP_j^{-1})$.

\begin{figure}[t!]
    \centering
    \includegraphics[width=\linewidth]{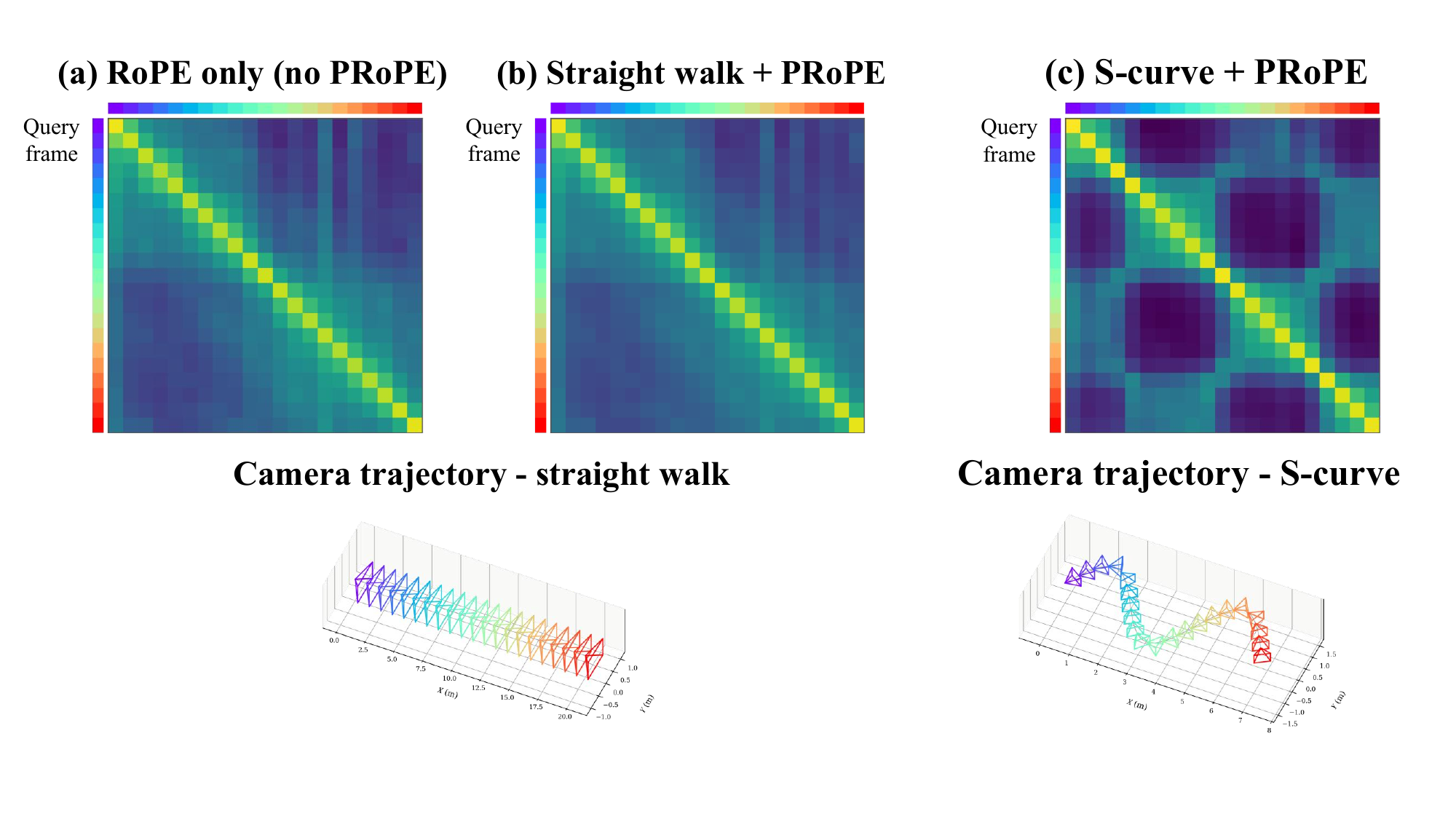}
    \caption{Tiled PRoPE and the native RoPE act together in one attention kernel: temporal
structure is preserved, and pose similarity is rewarded. We use two synthetic trajectories here to illustrate the preservation (bottom row), and calculate frame-by-frame attention
logits for heatmap visualization (top row). (a)~RoPE only (camera term off): the purely
temporal kernel, banded in frame offset. (b)~Straight walk with tiled PRoPE: the map is nearly indistinguishable from (a) --- the maximum change
is empirically 1.3\% of (a)'s dynamic range --- adding the camera does not disrupt the
frame-to-frame attention structure. (c)~S-curve with tiled PRoPE: frames
with similar heading light up in blocks far from the diagonal, and
opposite-heading frames darken. The camera raises attention exactly where
views are geometrically close, while the temporal ordering of (a) survives
untouched. This analysis was performed on the model fine-tuned from Wan2.2-TI2V-5B.}
    \label{fig:rope-intact}
\end{figure}

We quantify the interaction with a matched-content probe: two frames are assigned identical content, and we read the attention score they exchange, so that any difference in the score is attributable purely to the positional and camera encodings rather than to image content. Figure~\ref{fig:rope-intact} shows the resulting attention kernels on two controlled trajectories. Tiled PRoPE leaves the temporal structure of the RoPE-only kernel essentially untouched, and modulates attention only through relative viewpoint --- raising it between frames that see the scene from similar directions, lowering it between frames that do not. The temporal code is thus re-weighted by viewpoint, not overwritten.

\section{Camera Collision Avoidance via a Progressive 3D Occupancy Map}
\label{app:collision-avoidance}

\begin{figure}[t!]
    \centering
    \includegraphics[width=\linewidth]{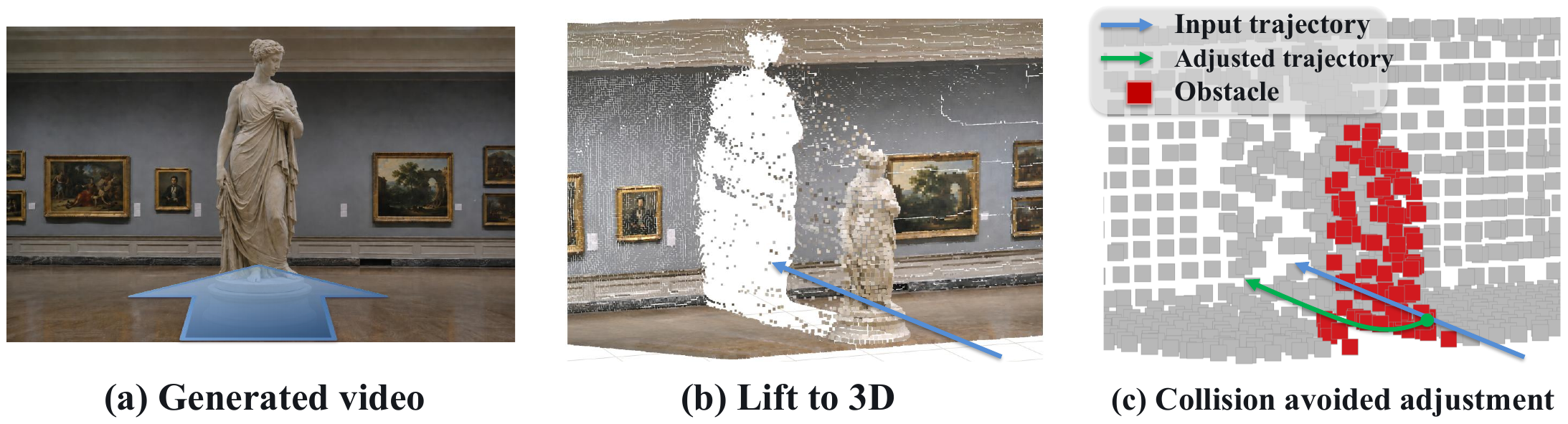}
    \caption{\textbf{Camera collision avoidance via a progressive 3D occupancy
    map.} (a) The generated frames of a rollout segment, with the planned input
    trajectory (blue) heading straight toward the statue. (b) Metric depth and
    camera pose lift the observed surfaces into a unified 3D world frame and
    aggregate them into voxels. (c) Voxels with enough surface observations are
    marked as occupied (red); the planned trajectory (blue) that would pass into
    this occupied region is corrected into an adjusted trajectory (green) that
    advances laterally along the scene structure while staying outside a preset
    safety margin.}
    \label{fig:collision-system}
\end{figure}

Matrix-Game 3.5 uses the camera trajectory as its generation control signal ---
during interaction, keyboard and mouse inputs are mapped to camera trajectories ---
so given a sequence of camera poses, the model faithfully advances along that
trajectory, while the control signal itself does not reason about collisions
between the trajectory and the scene structure. When a planned trajectory passes
into an already-generated wall, object, or other scene surface, the model must
still keep generating the content behind the obstacle and complete new scene
structure for that region. Because this newly generated structure is only weakly
constrained by the geometric surfaces observed so far, it is prone to 3D
inconsistency; such inconsistency further accumulates segment by segment over the
subsequent rollout and can eventually cause the generation to collapse. This
failure mode is specific to trajectories that penetrate scene surfaces: when the
camera explores free space, newly generated content remains well constrained by
the geometry observed so far. For
colliding trajectories, we provide an optional camera collision-avoidance mechanism
based on a progressive 3D occupancy map, so that camera motion can adapt to the
scene structure that has already been generated.

At the beginning of each rollout segment, the system estimates metric depth and camera
pose for the anchor (or for the generated frames of the previous segment),
back-projects the depth pixels into a unified world coordinate frame, and
aggregates them into 3D voxels according to their spatial positions; once a voxel
accumulates enough surface observations, it is marked as occupied. Segment~0
initializes the occupancy map from the anchor, and each subsequent segment keeps
adding geometric observations from previously generated content, so that the map
expands progressively as the rollout proceeds and forms a cumulative
representation of the observed scene structure.

Building on this cumulative scene representation, we apply a frame-by-frame
geometric correction to the originally planned camera trajectory before
generating each segment. The planned trajectory is first aligned to the
registered frame in which the occupancy map lives. Then, starting from the
corrected camera center of the previous frame, the current displacement is
decomposed into a radial component pointing toward the nearby occupied region and
a tangential component moving along its boundary. The correction suppresses the
radial component that would drive the camera further toward the obstacle while
preserving the tangential component and any component that moves away from the
obstacle; when a candidate camera center enters a preset safety margin, it is
further pushed back to the safety boundary along the direction away from the
occupied region. In this way, camera motion that originally headed straight into a
scene surface is turned into a trajectory that advances laterally along the scene
structure. The whole procedure adjusts only the camera position, keeps the
original rotation and camera intrinsics unchanged, and recomputes the translation
term of the world-to-camera extrinsics from the new camera center.

The corrected trajectory is used for both the patch-memory gather and tiled PRoPE, keeping the
geometric reprojection of the historical frames, the camera-position
conditioning, and the current generation trajectory mutually consistent. Once the
current segment has been generated, its frames are re-registered with depth and
pose estimation in the next segment and continue to update the 3D occupancy map,
forming a closed loop of scene registration, occupancy update, trajectory
correction, and video generation (Figure~\ref{fig:collision-system}). This method
lets camera motion adaptively adjust using the scene geometry that keeps
accumulating during rollout, while maintaining spatial consistency between the
patch memory and the camera conditioning.

\FloatBarrier

{
\small
\bibliography{neurips_2025}

@String(siggraph={ACM SIGGRAPH})

@String(wacv={WACV})

@article{zhu2026causal,
  title={Causal Forcing: Autoregressive Diffusion Distillation Done Right for High-Quality Real-Time Interactive Video Generation},
  author={Zhu, Hongzhou and Zhao, Min and He, Guande and Su, Hang and Li, Chongxuan and Zhu, Jun},
  journal={arXiv preprint arXiv:2602.02214},
  year={2026}
}

@inproceedings{bruce2024genie,
    author = {Bruce, Jake and Dennis, Michael D and Edwards, Ashley and Parker-Holder, Jack and Shi, Yuge and Hughes, Edward and Lai, Matthew and Mavalankar, Aditi and Steigerwald, Richie and Apps, Chris and others},
    booktitle = {International Conference on Machine Learning},
    title = {Genie: Generative interactive environments},
    year = {2024}
}

@misc{li2026distillalign,
  title         = {DistillAlign: Coordinating Mode Covering and Mode Seeking in Autoregressive Video Distillation},
  author        = {Li, Jiaxing and Zou, Kai and Zhou, Cindy and Huang, Kaichen and Gao, Junyao and Wang, Zile and Liu, Yang and Liu, Bin and An, Bo and Li, Yangguang},
  year          = {2026},
  eprint        = {2607.26811},
  archivePrefix = {arXiv},
  primaryClass  = {cs.CV},
  url           = {https://arxiv.org/abs/2607.26811}
}

@article{che2024gamegen,
    author = {Che, Haoxuan and He, Xuanhua and Liu, Quande and Jin, Cheng and Chen, Hao},
    journal = {arXiv preprint arXiv:2411.00769},
    title = {Gamegen-x: Interactive open-world game video generation},
    year = {2024}
}

@inproceedings{henschel2025streamingt2v,
  title={Streamingt2v: Consistent, dynamic, and extendable long video generation from text},
  author={Henschel, Roberto and Khachatryan, Levon and Poghosyan, Hayk and Hayrapetyan, Daniil and Tadevosyan, Vahram and Wang, Zhangyang and Navasardyan, Shant and Shi, Humphrey},
  booktitle={Proceedings of the Computer Vision and Pattern Recognition Conference},
  pages={2568--2577},
  year={2025}
}

@article{HaCohen2024LTXVideo,
    author = {HaCohen, Yoav and Chiprut, Nisan and Brazowski, Benny and Shalem, Daniel and Moshe, Dudu and Richardson, Eitan and Levin, Eran and Shiran, Guy and Zabari, Nir and Gordon, Ori and Panet, Poriya and Weissbuch, Sapir and Kulikov, Victor and Bitterman, Yaki and Melumian, Zeev and Bibi, Ofir},
    journal = {arXiv preprint arXiv:2501.00103},
    title = {LTX-Video: Realtime Video Latent Diffusion},
    year = {2024}
}

@article{yin2024improved,
  title={Improved distribution matching distillation for fast image synthesis},
  author={Yin, Tianwei and Gharbi, Micha{\"e}l and Park, Taesung and Zhang, Richard and Shechtman, Eli and Durand, Fredo and Freeman, Bill},
  journal={Advances in neural information processing systems},
  volume={37},
  pages={47455--47487},
  year={2024}
}

@inproceedings{yin2024one,
  title={One-step diffusion with distribution matching distillation},
  author={Yin, Tianwei and Gharbi, Micha{\"e}l and Zhang, Richard and Shechtman, Eli and Durand, Fredo and Freeman, William T and Park, Taesung},
  booktitle={Proceedings of the IEEE/CVF conference on computer vision and pattern recognition},
  pages={6613--6623},
  year={2024}
}

@article{he2024cameractrl,
    author = {He, Hao and Xu, Yinghao and Guo, Yuwei and Wetzstein, Gordon and Dai, Bo and Li, Hongsheng and Yang, Ceyuan},
    journal = {arXiv preprint arXiv:2404.02101},
    title = {Cameractrl: Enabling camera control for text-to-video generation},
    year = {2024}
}

@inproceedings{huang2024vbench,
    author = {Huang, Ziqi and He, Yinan and Yu, Jiashuo and Zhang, Fan and Si, Chenyang and Jiang, Yuming and Zhang, Yuanhan and Wu, Tianxing and Jin, Qingyang and Chanpaisit, Nattapol and others},
    booktitle = {Computer Vision and Pattern Recognition},
    pages = {21807--21818},
    title = {Vbench: Comprehensive benchmark suite for video generative models},
    year = {2024}
}

@article{huang2025self,
    author = {Huang, Xun and Li, Zhengqi and He, Guande and Zhou, Mingyuan and Shechtman, Eli},
    journal = {arXiv preprint arXiv:2506.08009},
    title = {Self Forcing: Bridging the Train-Test Gap in Autoregressive Video Diffusion},
    year = {2025}
}

@article{yu2026mosaicmem,
  title={MosaicMem: Hybrid Spatial Memory for Controllable Video World Models},
  author={Yu, Wei and Qian, Runjia and Li, Yumeng and Wang, Liquan and Yin, Songheng and Anthony, Dennis and Ye, Yang and Li, Yidi and Wan, Weiwei and Garg, Animesh and others},
  journal={arXiv preprint arXiv:2603.17117},
  year={2026}
}

@article{wan2025wan,
  title={Wan: Open and advanced large-scale video generative models},
  author={Wan, Team and Wang, Ang and Ai, Baole and Wen, Bin and Mao, Chaojie and Xie, Chen-Wei and Chen, Di and Yu, Feiwu and Zhao, Haiming and Yang, Jianxiao and others},
  journal={arXiv preprint arXiv:2503.20314},
  year={2025}
}

@article{kong2024hunyuanvideo,
    author = {Kong, Weijie and Tian, Qi and Zhang, Zijian and Min, Rox and Dai, Zuozhuo and Zhou, Jin and Xiong, Jiangfeng and Li, Xin and Wu, Bo and Zhang, Jianwei and others},
    journal = {arXiv preprint arXiv:2412.03603},
    title = {Hunyuanvideo: A systematic framework for large video generative models},
    year = {2024}
}

@article{oasis2024,
    author = {Decart},
    title = {Oasis: A Universe in a Transformer},
    url = {https://oasis-model.github.io/},
    year = {2024}
}

@misc{openai2024worldsim,
    author = {{OpenAI}},
    howpublished = {\url{https://openai.com/index/video-generation-models-as-world-simulators/}},
    title = {Sora: Video Generation Models as World Simulators},
    year = {2024}
}

@inproceedings{peebles2023scalable,
    author = {Peebles, William and Xie, Saining},
    booktitle = {International Conference on Computer Vision},
    pages = {4195--4205},
    title = {Scalable diffusion models with transformers},
    year = {2023}
}

@inproceedings{rombach2022high,
    author = {Rombach, Robin and Blattmann, Andreas and Lorenz, Dominik and Esser, Patrick and Ommer, Bj{\"o}rn},
    booktitle = {Computer Vision and Pattern Recognition},
    pages = {10684--10695},
    title = {High-resolution image synthesis with latent diffusion models},
    year = {2022}
}

@article{rope,
  author       = {Jianlin Su and
                  Yu Lu and
                  Shengfeng Pan and
                  Bo Wen and
                  Yunfeng Liu},
  title        = {RoFormer: Enhanced Transformer with Rotary Position Embedding},
  journal      = {CoRR},
  volume       = {abs/2104.09864},
  year         = {2021},
  url          = {https://arxiv.org/abs/2104.09864},
  eprinttype    = {arXiv},
  eprint       = {2104.09864},
  bibsource    = {dblp computer science bibliography, https://dblp.org}
}

@article{xiao2025worldmem,
    author = {Xiao, Zeqi and Lan, Yushi and Zhou, Yifan and Ouyang, Wenqi and Yang, Shuai and Zeng, Yanhong and Pan, Xingang},
    journal = {arXiv preprint arXiv:2504.12369},
    title = {WORLDMEM: Long-term Consistent World Simulation with Memory},
    year = {2025}
}

@inproceedings{yin2025slow,
    author = {Yin, Tianwei and Zhang, Qiang and Zhang, Richard and Freeman, William T and Durand, Fredo and Shechtman, Eli and Huang, Xun},
    booktitle = {Proceedings of the Computer Vision and Pattern Recognition Conference},
    pages = {22963--22974},
    title = {From slow bidirectional to fast autoregressive video diffusion models},
    year = {2025}
}

@inproceedings{yu2025gamefactory,
    author = {Yu, Jiwen and Qin, Yiran and Wang, Xintao and Wan, Pengfei and Zhang, Di and Liu, Xihui},
    booktitle = {International Conference on Computer Vision},
    title = {GameFactory: Creating New Games with Generative Interactive Videos},
    year = {2025}
}

@article{teng2025magi,
  title={Magi-1: Autoregressive video generation at scale},
  author={Teng, Hansi and Jia, Hongyu and Sun, Lei and Li, Lingzhi and Li, Maolin and Tang, Mingqiu and Han, Shuai and Zhang, Tianning and Zhang, WQ and Luo, Weifeng and others},
  journal={arXiv preprint arXiv:2505.13211},
  year={2025}
}

@article{cui2025self,
  title={Self-forcing++: Towards minute-scale high-quality video generation},
  author={Cui, Justin and Wu, Jie and Li, Ming and Yang, Tao and Li, Xiaojie and Wang, Rui and Bai, Andrew and Ban, Yuanhao and Hsieh, Cho-Jui},
  journal={arXiv preprint arXiv:2510.02283},
  year={2025}
}

@inproceedings{yu2025context,
  title={Context as memory: Scene-consistent interactive long video generation with memory retrieval},
  author={Yu, Jiwen and Bai, Jianhong and Qin, Yiran and Liu, Quande and Wang, Xintao and Wan, Pengfei and Zhang, Di and Liu, Xihui},
  booktitle={Proceedings of the SIGGRAPH Asia 2025 Conference Papers},
  pages={1--11},
  year={2025}
}

@inproceedings{li2025vmem,
  title={Vmem: Consistent interactive video scene generation with surfel-indexed view memory},
  author={Li, Runjia and Torr, Philip and Vedaldi, Andrea and Jakab, Tomas},
  booktitle={Proceedings of the IEEE/CVF International Conference on Computer Vision},
  pages={25690--25699},
  year={2025}
}

@article{he2025matrix,
  title={Matrix-game 2.0: An open-source real-time and streaming interactive world model},
  author={He, Xianglong and Peng, Chunli and Liu, Zexiang and Wang, Boyang and Zhang, Yifan and Cui, Qi and Kang, Fei and Jiang, Biao and An, Mengyin and Ren, Yangyang and others},
  journal={arXiv preprint arXiv:2508.13009},
  year={2025}
}

@article{hong2025relic,
  title={Relic: Interactive video world model with long-horizon memory},
  author={Hong, Yicong and Mei, Yiqun and Ge, Chongjian and Xu, Yiran and Zhou, Yang and Bi, Sai and Hold-Geoffroy, Yannick and Roberts, Mike and Fisher, Matthew and Shechtman, Eli and others},
  journal={arXiv preprint arXiv:2512.04040},
  year={2025}
}

@article{sun2025worldplay,
  title={Worldplay: Towards long-term geometric consistency for real-time interactive world modeling},
  author={Sun, Wenqiang and Zhang, Haiyu and Wang, Haoyuan and Wu, Junta and Wang, Zehan and Wang, Zhenwei and Wang, Yunhong and Zhang, Jun and Wang, Tengfei and Guo, Chunchao},
  journal={arXiv preprint arXiv:2512.14614},
  year={2025}
}

@article{parkerholder2024genie2,
  title         = {Genie 2: A Large-Scale Foundation World Model},
  author        = {Jack Parker-Holder and Philip Ball and Jake Bruce and Vibhavari Dasagi and Kristian Holsheimer and Christos Kaplanis and Alexandre Moufarek and Guy Scully and Jeremy Shar and Jimmy Shi and Stephen Spencer and Jessica Yung and Michael Dennis and Sultan Kenjeyev and Shangbang Long and Vlad Mnih and Harris Chan and Maxime Gazeau and Bonnie Li and Fabio Pardo and Luyu Wang and Lei Zhang and Frederic Besse and Tim Harley and Anna Mitenkova and Jane Wang and Jeff Clune and Demis Hassabis and Raia Hadsell and Adrian Bolton and Satinder Singh and Tim Rockt{\"a}schel},
  year          = {2024},
  url           = {https://deepmind.google/discover/blog/genie-2-a-large-scale-foundation-world-model/}
}

@article{team2026advancing,
  title={Advancing Open-source World Models},
  author={Team, Robbyant and Gao, Zelin and Wang, Qiuyu and Zeng, Yanhong and Zhu, Jiapeng and Cheng, Ka Leong and Li, Yixuan and Wang, Hanlin and Xu, Yinghao and Ma, Shuailei and others},
  journal={arXiv preprint arXiv:2601.20540},
  year={2026}
}

@article{zhu2026sana,
  title={Sana-wm: Efficient minute-scale world modeling with hybrid linear diffusion transformer},
  author={Zhu, Haoyi and Liu, Haozhe and Zhao, Yuyang and Ye, Tian and Chen, Junsong and Yu, Jincheng and He, Tong and Han, Song and Xie, Enze},
  journal={arXiv preprint arXiv:2605.15178},
  year={2026}
}

@article{zhao2026perceptual,
  title={Perceptual Flow Matching for Few-Step Generative Modeling},
  author={Zhao, Chuyang and Song, Yifei and Wang, Hongfa and Yuan, Jianlong and Zhang, Yuan and Fu, Siming and Chen, Zhineng and Deng, Huilin and Huang, Haoyang and Duan, Nan},
  journal={arXiv preprint arXiv:2607.03524},
  year={2026}
}

@article{wang2024internvideo2,
  title={InternVideo2: Scaling Foundation Models for Multimodal Video Understanding},
  author={Wang, Yi and Li, Kunchang and Li, Xinhao and Yu, Jiashuo and He, Yinan and Wang, Chenting and Chen, Guo and Pei, Baoqi and Yan, Ziang and Zheng, Rongkun and Xu, Jilan and Wang, Zun and Shi, Yansong and Jiang, Tianxiang and Li, Songze and Zhang, Hongjie and Huang, Yifei and Qiao, Yu and Wang, Yali and Wang, Limin},
  journal={arXiv preprint arXiv:2403.15377},
  year={2024}
}

@article{zou2026hiar,
  title={HiAR: Efficient Autoregressive Long Video Generation via Hierarchical Denoising},
  author={Zou, Kai and Zheng, Dian and Liu, Hongbo and Hang, Tiankai and Liu, Bin and Yu, Nenghai},
  journal={arXiv preprint arXiv:2603.08703},
  year={2026}
}

@article{wang2026matrixgame3,
  title={Matrix-game 3.0: Real-time and streaming interactive  world model with long-horizon memory},
  author={Wang, Zile and Liu, Zexiang and Li, Jiaxing and Huang, Kaichen and Xu, Baixin and Kang, Fei and An, Mengyin and Wang, Peiyu and Jiang, Biao and Wei, Yichen and others},
  journal={arXiv preprint arXiv:2604.08995},
  year={2026}
}

@article{gao2026infiniteworlds,
  title={Infinite Worlds with Versatile Interactions},
  author={Gao, Zelin and Wang, Qiuyu and Zhu, Jiapeng and Chen, Jingye and Liu, Zichen and Bai, Qingyan and Wang, Jiahao and Yuan, Yufeng and Wang, Hanlin and Lu, Yichong and others},
  journal={arXiv preprint arXiv:2607.07534},
  year={2026}
}

@article{li2025prope,
  title={Cameras as relative positional encoding},
  author={Li, Ruilong and Yi, Brent and Liu, Junchen and Gao, Hang and Ma, Yi and Kanazawa, Angjoo},
  journal={Advances in Neural Information Processing Systems},
  volume={38},
  pages={15984--16009},
  year={2026}
}

@article{wang2024boximator,
  title={Boximator: Generating rich and controllable motions for video synthesis},
  author={Wang, Jiawei and Zhang, Yuchen and Zou, Jiaxin and Zeng, Yan and Wei, Guoqiang and Yuan, Liping and Li, Hang},
  journal={arXiv preprint arXiv:2402.01566},
  year={2024}
}

@inproceedings{li2025trackdiffusion,
  title={Trackdiffusion: Tracklet-conditioned video generation via diffusion models},
  author={Li, Pengxiang and Chen, Kai and Liu, Zhili and Gao, Ruiyuan and Hong, Lanqing and Yeung, Dit-Yan and Lu, Huchuan and Jia, Xu},
  booktitle={2025 IEEE/CVF Winter Conference on Applications of Computer Vision (WACV)},
  pages={3539--3548},
  year={2025},
  organization={IEEE}
}

@inproceedings{zhang2026tgt,
  title={Tgt: Text-grounded trajectories for locally controlled video generation},
  author={Zhang, Guofeng and Wang, Angtian and Fang, Jacob Zhiyuan and Jiang, Liming and Yang, Haotian and Liu, Bo and Yang, Yiding and Chen, Guang and Wen, Longyin and Yuille, Alan and others},
  booktitle={Proceedings of the IEEE/CVF Conference on Computer Vision and Pattern Recognition},
  pages={22028--22037},
  year={2026}
}

@article{wei2024dreamvideo,
  title={Dreamvideo-2: Zero-shot subject-driven video customization with precise motion control},
  author={Wei, Yujie and Zhang, Shiwei and Yuan, Hangjie and Wang, Xiang and Qiu, Haonan and Zhao, Rui and Feng, Yutong and Liu, Feng and Huang, Zhizhong and Ye, Jiaxin and others},
  journal={arXiv preprint arXiv:2410.13830},
  year={2024}
}

@article{mai2025contextanyone,
  title={ContextAnyone: Context-Aware Diffusion for Character-Consistent Text-to-Video Generation},
  author={Mai, Ziyang and Tai, Yu-Wing},
  journal={arXiv preprint arXiv:2512.07328},
  year={2025}
}

@article{wei2026memento,
  title={Memento: Reconstruct to Remember for Consistent Long Video Generation},
  author={Wei, Xuan and Ji, Longbin and Wang, Guan and Liu, Xiangrui and Zhang, Zhenyu and Wang, Shuohuan and Sun, Yu and Hong, Qingqi},
  journal={arXiv preprint arXiv:2606.14667},
  year={2026}
}

@article{akdemir2026zero,
  title={From Zero to Hero: Training-Free Custom Concept Spawning in World Models},
  author={Akdemir, Kiymet and Yanardag, Pinar},
  journal={arXiv preprint arXiv:2606.02575},
  year={2026}
}

@article{wang2026worlddirector,
  title={WorldDirector: Building Controllable World Simulators with Persistent Dynamic Memory},
  author={Wang, Hanlin and Ouyang, Hao and Wang, Qiuyu and Wang, Wen and Bai, Qingyan and Cheng, Ka Leong and Yu, Yue and Li, Yixuan and Meng, Yihao and Liu, Zichen and others},
  journal={arXiv preprint arXiv:2607.02517},
  year={2026}
}

@article{xiong2026actworld,
  title={ActWorld: From Explorable to Interactive World Model via Action-Aware Memory},
  author={Xiong, Zhexiao and Song, Yizhi and Kang, Hao and Yan, Qing and Jiang, Liming and Yang, Jenson and Fu, Zhoujie and Fotiadis, Stathi and Wang, Angtian and Liu, Zichuan and others},
  journal={arXiv preprint arXiv:2606.17730},
  year={2026}
}

@article{wang2026scenecode,
  title={SceneCode: Executable World Programs for Editable Indoor Scenes with Articulated Objects},
  author={Wang, Puyi and Wang, Yuhao and Li, Linjie and Yang, Zhengyuan and Lin, Kevin Qinghong and Li, Yangguang and Cheng, Yu},
  journal={arXiv preprint arXiv:2605.19587},
  year={2026}
}

@article{zhao2026causalforcingplusplus,
  title={Causal forcing++: Scalable few-step autoregressive diffusion distillation for real-time interactive video generation},
  author={Zhao, Min and Zhu, Hongzhou and Zheng, Kaiwen and Zhou, Zihan and Yan, Bokai and Li, Xinyuan and Yang, Xiao and Li, Chongxuan and Zhu, Jun},
  journal={arXiv preprint arXiv:2605.15141},
  year={2026}
}

@inproceedings{zhuang2025flashvsr,
  title={FlashVSR: Towards Real-time Diffusion-Based Streaming Video Super Resolution},
  author={Zhuang, Junhao and Guo, Shi and Cai, Xin and Li, Xiaohui and Liu, Yihao and Yuan, Chun and Xue, Tianfan},
  booktitle={Proceedings of the IEEE/CVF Conference on Computer Vision and Pattern Recognition},
  pages={43482--43493},
  year={2026}
}

@article{wang2026vggtomega,
  title={{VGGT-$\Omega$}},
  author={Wang, Jianyuan and Chen, Minghao and Zhang, Shangzhan and Karaev, Nikita and Sch{\"o}nberger, Johannes and Labatut, Patrick and Bojanowski, Piotr and Novotny, David and Vedaldi, Andrea and Rupprecht, Christian},
  journal={arXiv preprint arXiv:2605.15195},
  year={2026}
}

@article{lin2025depthanything3,
  title={Depth anything 3: Recovering the visual space from any views},
  author={Lin, Haotong and Chen, Sili and Liew, Junhao and Chen, Donny Y and Li, Zhenyu and Shi, Guang and Feng, Jiashi and Kang, Bingyi},
  journal={arXiv preprint arXiv:2511.10647},
  year={2025}
}

@article{team2026gemma,
  title={Gemma 4 technical report},
  author={Team, Gemma and Abd, Sherif El and Aggarwal, Vaibhav and Algayres, Robin and Andreev, Alek and Bachem, Olivier and Ballantyne, Ian and Brick, Cormac and C{\u{a}}rbune, Victor and Casbon, Michelle and others},
  journal={arXiv preprint arXiv:2607.02770},
  year={2026}
}

@article{sapkota2025ultralytics,
  title={Ultralytics YOLO evolution: An overview of YOLO26, YOLO11, YOLOv8 and YOLOv5 object detectors for computer vision and pattern recognition},
  author={Sapkota, Ranjan and Karkee, Manoj},
  journal={arXiv preprint arXiv:2510.09653},
  year={2025}
}

@misc{simeoni2025dinov3,
  title={{DINOv3}},
  author={Sim{\'e}oni, Oriane and Vo, Huy V. and Seitzer, Maximilian and Baldassarre, Federico and Oquab, Maxime and Jose, Cijo and Khalidov, Vasil and Szafraniec, Marc and Yi, Seungeun and Ramamonjisoa, Micha{\"e}l and Massa, Francisco and Haziza, Daniel and Wehrstedt, Luca and Wang, Jianyuan and Darcet, Timoth{\'e}e and Moutakanni, Th{\'e}o and Sentana, Leonel and Roberts, Claire and Vedaldi, Andrea and Tolan, Jamie and Brandt, John and Couprie, Camille and Mairal, Julien and J{\'e}gou, Herv{\'e} and Labatut, Patrick and Bojanowski, Piotr},
  year={2025},
  eprint={2508.10104},
  archivePrefix={arXiv},
  primaryClass={cs.CV},
  url={https://arxiv.org/abs/2508.10104},
}

@article{wang2004ssim,
  title={Image quality assessment: from error visibility to structural similarity},
  author={Wang, Zhou and Bovik, Alan C and Sheikh, Hamid R and Simoncelli, Eero P},
  journal={IEEE transactions on image processing},
  volume={13},
  number={4},
  pages={600--612},
  year={2004},
  publisher={IEEE}
}

@inproceedings{zhang2018lpips,
  title={The unreasonable effectiveness of deep features as a perceptual metric},
  author={Zhang, Richard and Isola, Phillip and Efros, Alexei A and Shechtman, Eli and Wang, Oliver},
  booktitle={Proceedings of the IEEE conference on computer vision and pattern recognition},
  pages={586--595},
  year={2018}
}

@article{chen2025deepverse,
  title={Deepverse: 4d autoregressive video generation as a world model},
  author={Chen, Junyi and Zhu, Haoyi and He, Xianglong and Wang, Yifan and Zhou, Jianjun and Chang, Wenzheng and Zhou, Yang and Li, Zizun and Fu, Zhoujie and Pang, Jiangmiao and others},
  journal={arXiv preprint arXiv:2506.01103},
  year={2025}
}

@article{dreamx2026world,
  title={DreamX-World 1.0: A General-Purpose Interactive World Model},
  author={{DreamX Team} and Bai, Yancheng and Chen, Rui and Chu, Xiangxiang and others},
  journal={arXiv preprint arXiv:2606.16993},
  year={2026}
}

@article{bai2025pefield,
  title={Positional Encoding Field},
  author={Bai, Yunpeng and Li, Haoxiang and Huang, Qixing},
  journal={arXiv preprint arXiv:2510.20385},
  year={2025}
}

@article{zhuang2026self,
  title={Self Gradient Forcing: Native Long Video Extrapolation},
  author={Zhuang, Junhao and Zhang, Shiyi and Bian, Yuxuan and Li, Yaowei and Luo, Yawen and Liu, Yijun and Jin, Weiyang and Zhang, Songchun and He, Xianglong and Zhang, Xuying and others},
  journal={arXiv preprint arXiv:2607.20368},
  year={2026}
}
\bibliographystyle{unsrt}
}

\end{document}